\documentclass[runningheads]{llncs}

\usepackage{eccv}

\usepackage{eccvabbrv}

\usepackage{graphicx}
\usepackage{booktabs}

\usepackage[accsupp]{axessibility}  

\usepackage{cite}
\usepackage{mathrsfs}
\usepackage{amsfonts}
\usepackage{amsmath}
\usepackage{amssymb}
\usepackage{bbding} 
\usepackage{booktabs}
\usepackage{tabularx}
\usepackage{multirow}
\usepackage{xcolor}
\usepackage{algorithm}

\usepackage{mathtools}
\usepackage{algorithm}
\usepackage[noEnd,indLines]{algpseudocodex}

\usepackage[abs]{overpic}
\newcommand{\tabincell}[2]{\begin{tabular}{@{}#1@{}}#2\end{tabular}}

\def\eg{\emph{e.g}\onedot} 
\def\ie{\emph{i.e}\onedot} 

\usepackage[pagebackref,breaklinks,colorlinks]{hyperref}

\usepackage{orcidlink}

\begin{document}


\title{Dual-Camera Smooth Zoom on Mobile Phones}   

\author{Renlong Wu 
\and
Zhilu Zhang$^{(}$\Envelope$^)$ 
\and
Yu Yang
\and
Wangmeng Zuo 
}

\authorrunning{R.Wu et al.}

\institute{
Harbin Institute of Technology, China
\\
\email{hirenlongwu@gmail.com,cszlzhang@outlook.com,\\806224005qq@gmail.com,wmzuo@hit.edu.cn}
\\
Project: \url{https://dualcamerasmoothzoom.github.io}
}

\maketitle

\begin{abstract}
 When zooming between dual cameras on a mobile, noticeable jumps in geometric content and image color occur in the preview, inevitably affecting the user’s zoom experience. In this work, we introduce a new task, \ie, dual-camera smooth zoom (DCSZ) to achieve a smooth zoom preview. The frame interpolation (FI) technique is a potential solution but struggles with ground-truth collection. To address the issue, we suggest a data factory solution where continuous virtual cameras are assembled to generate DCSZ data by rendering  reconstructed 3D models of the scene. In particular, we propose a novel dual-camera smooth zoom Gaussian Splatting (ZoomGS), where a camera-specific encoding is introduced to construct a specific 3D model for each virtual camera. With the proposed data factory, we construct a synthetic dataset for DCSZ, and we utilize it to fine-tune FI models. In addition, we collect real-world dual-zoom images without ground-truth for evaluation. Extensive experiments are conducted with multiple FI methods. The results show that the fine-tuned FI models achieve a significant performance improvement over the original ones on DCSZ task. The datasets, codes, and pre-trained models are available at \url{https://github.com/ZcsrenlongZ/ZoomGS}.
 

\end{abstract}

\section{Introduction}
\begin{figure}[t!]
    \centering
    \includegraphics[width=0.99\linewidth]{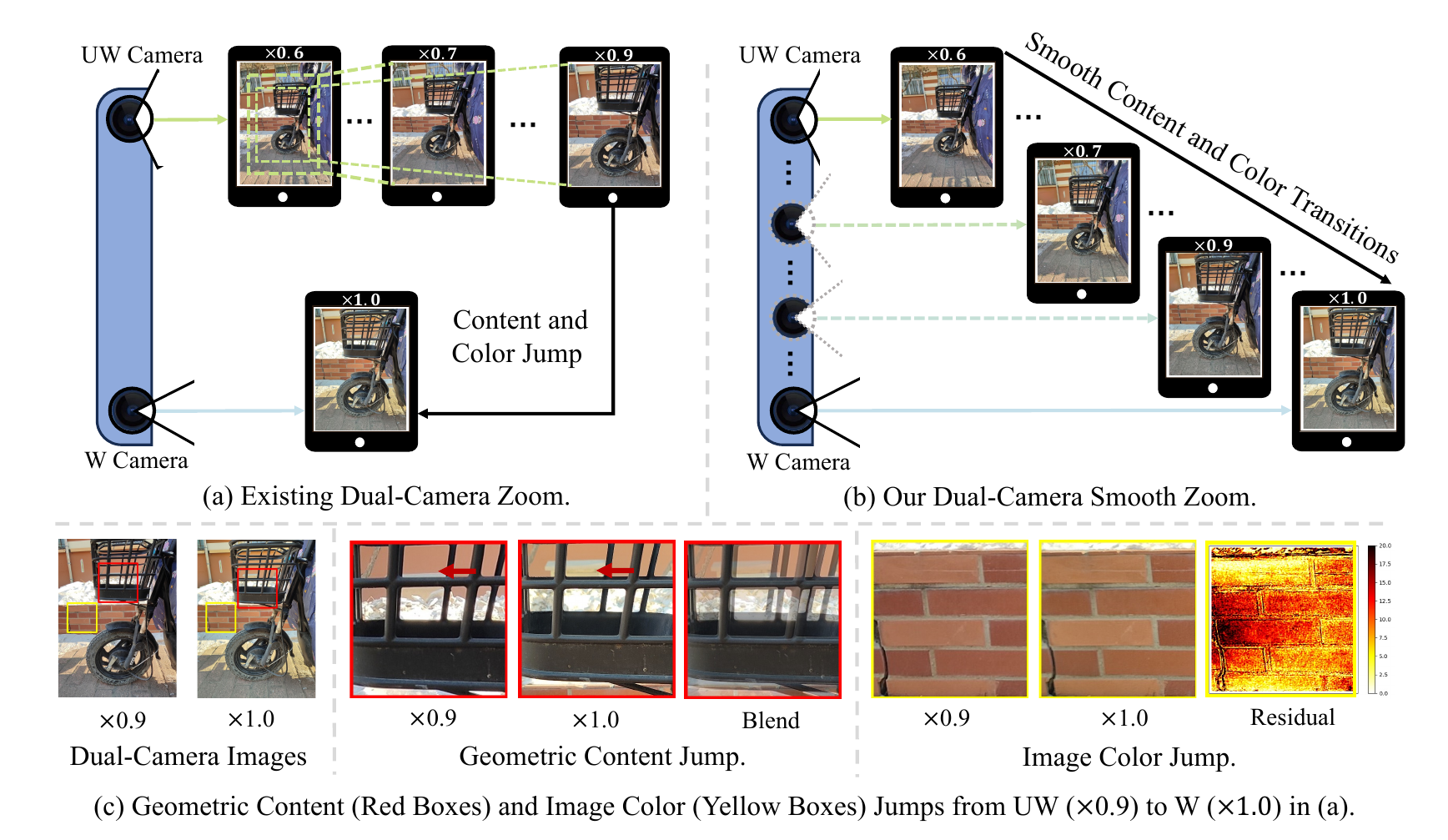}
    \caption{
    (a) Existing dual-camera zoom.
    For zooming between UW and W cameras (\ie, from $\times 0.6$ to $\times 1.0$), smartphones (\eg, Xiaomi, OPPO, and vivo) generally crop out the specific area from the UW image, and scale the image up to the dimensions of the original.
    When the zoom factor changes from 0.9 to 1.0, the lens has to switch from UW to W, where notable geometric content and image color jump happen in the preview.
    (b) Our proposed dual-camera smooth zoom (DCSZ).
    (c) The geometric content and image color jump in (a) existing dual-camera zoom. 
    Some examples can be found \href{https://dualcamerasmoothzoom.github.io}{here}.
    }
    \label{fig:intro}
\end{figure}

Constrained by space, most mobile phones opt for multiple fixed-focal-length cameras for digital zoom, instead of a variable-focal-length lens for optical zoom~\cite{blahnik2021smartphone}.
In particular, dual cameras have been commonly equipped, consisting of a wide-angle (\ie, $\times 1.0$) lens for primary imaging and an ultra-wide-angle (\ie, $\times 0.5$ or $\times 0.6$) lens for an even wider view.
For zooming between the ultra-wide-angle (UW) and wide-angle (W) cameras, the current smartphones (\eg, Xiaomi, OPPO, and vivo) generally crop out the specific area from the UW image, and scale the image up to the dimensions of the original, as shown in \cref{fig:intro}(a).
However, when the zoom factor changes from 0.9 to 1.0, the lens has to switch from UW to W. 
The distinct deployment locations and internal characteristics~\cite{zhang2022self,lee2022reference,alzayer2023dc2,abdelhamed2021leveraging} of the two cameras lead to a notable preview jump in geometric content and image color (see \cref{fig:intro}(c)).
It significantly affects the user experience as users generally prefer smooth zoom.
To address the issue, we introduce a new task, \ie, dual-camera smooth zoom (DCSZ), aiming to achieve a fluid zoom preview on mobile phones for better zoom experience, as shown in \cref{fig:intro}(b).

Actually, frame interpolation (FI) techniques~\cite{EDSC,cheng2020video,lee2020adacof,niklaus2017video,niklaus2017video,peleg2019net,choi2020channel,kalluri2023flavr,kim2020fisr,shi2022video,bao2019depth,bao2019memc,RIFE,IFRNet,liu2017video,niklaus2018context,niklaus2020softmax,xu2019quadratic,xue2019video,UPRNet,AMT,EMAVFI,VFIFormer, BiFormer} can be a potential solution by synthesizing the intermediate images between dual-camera ones.
However, when applying existing pre-trained FI models, their results are unsatisfactory, due to the inevitable motion domain gap between the training and dual-camera data.
The more appropriate way is to exploit DCSZ data to retrain or fine-tune an FI model.
Note that the real-world ground-truth is hardly obtained, as it is impractical to deploy additional physical cameras between dual cameras.
Instead, we propose to assemble continuous virtual cameras to construct DCSZ data.

The virtual (V) camera must account for extrinsic and intrinsic parameters, as well as other imaging characteristics (\eg, image signal processing (ISP) pipeline).
Extrinsic and intrinsic parameters establish a mapping from 3D world coordinates to 2D image coordinates, determining the geometric content of the captured image.
Meanwhile, ISP affects the color rendition of the image.
Assuming all parameters for UW and W cameras are known, the parameters of a V camera can be obtained by interpolating the two. 
Thus, generating an image captured by the V camera only requires obtaining the 3D model of the scene.
Unfortunately, some parameters such as ISP are generally not available and their interpolation may be too complex.
To circumvent the acquisition of these parameters, we propose to construct specific 3D models for each V camera instead of a single camera-independent 3D model.
In this way, with the extrinsic and intrinsic parameters, the image to be captured by a V camera can be rendered from the 3D model of this V camera.

Specifically, we utilize recent 3DGS~\cite{3DGS} as our 3D representation manner and propose a novel dual-camera smooth zoom Gaussian Splatting (ZoomGS) method to achieve the above goal.
ZoomGS introduces a camera-specific encoding (\ie, $e$ in \cref{fig:pipeline}(b)) to decouple the geometric content of scene and the camera-dependent characteristics.
It takes the UW camera Gaussians as a base 3D model, and transforms it into the specific V camera ones according to $e$.

The overall pipeline of deploying virtual cameras is called a data factory.
With the proposed data factory, we construct a synthetic dataset for DCSZ, which is utilized to fine-tune FI models.
In addition, we collect real-world dual-camera images for evaluating FI models. 
Extensive experiments on multiple FI methods show that the fine-tuned models achieve a significant performance improvement over the original ones on DCSZ task both quantitatively and qualitatively, which demonstrates the effectiveness of the proposed data construction method.

The contributions can be summarized as follows:
\begin{itemize}
    \item We explore a practical task, \ie, dual-camera smooth zoom (DCSZ), and propose a solution that first constructs zoom sequences by a data factory and then utilizes them to fine-tune frame interpolation (FI) models.
    \item In the data factory, we propose a novel dual-camera smooth zoom Gaussian Splatting (ZoomGS) method, which decouples the geometric content of scene and the camera-dependent characteristics, achieving smooth zoom through continuous interpolation of extrinsic parameters, intrinsic parameters, and camera-specific encodings. 
    \item Beyond the synthetic data generated by the data factory, we also collect real-world dual-camera images for evaluation.
    Extensive experiments using multiple FI models demonstrate the effectiveness of the proposed method.

\end{itemize}

\section{Related Work}
\subsection{Frame Interpolation}
Given an image pair, frame interpolation (FI) aims to generate intermediate frames, which can be roughly divided into kernel-based methods~\cite{EDSC,cheng2020video,lee2020adacof,niklaus2017video,niklaus2017video,peleg2019net},
hallucination-based methods~\cite{choi2020channel,kalluri2023flavr,kim2020fisr,shi2022video}, and flow-based ones~\cite{bao2019depth,bao2019memc,RIFE,IFRNet,liu2017video,niklaus2018context,niklaus2020softmax,xu2019quadratic,xue2019video,UPRNet,AMT,EMAVFI,VFIFormer,BiFormer}.
Thanks to the robustness of optical flow, flow-based methods have become mainstream and are widely explored in optimization objectives~\cite{RIFE,IFRNet} and network designs~\cite{bao2019depth, bao2019memc, AMT,UPRNet, liu2017video,niklaus2018context, niklaus2020softmax, xu2019quadratic,EMAVFI,VFIFormer, BiFormer}.
For example, RIFE~\cite{RIFE} and IFRNet~\cite{IFRNet} incorporate task-oriented flow distillation objectives to generate intermediate flow prior during training.
UPRNet~\cite{UPRNet} exploits lightweight recurrent modules for both bi-directional
flow estimation and intermediate frame synthesis.
AMT~\cite{AMT} builds bidirectional correlation volumes for all pairs of pixels to retrieve correlations for more accurate flow estimation.
EMAVFI~\cite{EMAVFI} explicitly extracts motion and appearance information in a unifying way.
In addition, some works~\cite{VFIFormer, BiFormer} introduce vision transformer~\cite{dosovitskiy2020image} for performance improvement.
However, the unavailability of real-world ground-truth limits their effective applications for the DCSZ task.

\subsection{3D Reconstruction}

With the development of deep learning, neural rendering algorithms~\cite{NeRF,barron2021mip,martin2021nerf,yang2023freenerf,pumarola2021d,muller2022instant} have outperformed against traditional methods~\cite{szeliski1996lumigraph,hanrahan1996light} in 3D reconstruction.
Neural Radiance Field (NeRF)~\cite{NeRF} pioneers the track and facilitates high-quality novel view synthesis with the use of MLPs.
Advancements have been made to handle multi-resolution image inputs~\cite{barron2021mip}, sparse-view image inputs~\cite{yang2023freenerf}, dynamic scenes~\cite{pumarola2021d}, and nearly real-time rendering capabilities~\cite{muller2022instant}.
In addition, a few works~\cite{tosi2023nerf,tosi2023nerf,yen2022nerf} exploit the 3D priors in NeRF models to construct data for down-stream tasks, such as stereo depth estimation~\cite{tosi2023nerf}, object detection~\cite{ge2022neural}, and object dense description~\cite{yen2022nerf}.
However, the NeRF-based 3D reconstruction methods suffer from achieving photo-realistic rendering quality and real-time rendering speed simultaneously, due to the implicit scene representation and the dense sampling of rays.

In contrast, the recent 3D Gaussian Splatting (3DGS)~\cite{3DGS} overcomes the issue with explicit scene representation and differential rasterizer rendering, attracting lots of attention~\cite{chen2024survey}.
For example, some works utilize it for 3D reconstruction in sparse-view scenarios~\cite{chung2023depth,xiong2023sparsegs,yang2024gaussianobject,zhu2023fsgs,szymanowicz2023splatter,charatan2023pixelsplat}, dynamic scenarios~\cite{luiten2023dynamic,wu20234d,yang2023deformable,yang2023real,katsumata2023efficient,lin2023gaussian}, and COLMAP-free scenarios~\cite{fu2023colmap,keetha2023splatam,yan2023gs,huang2023photo}.
Some works employ it in segmentation~\cite{dou2024cosseggaussians,hu2024semantic,ye2023gaussian,cen2023segment,lan20232d}, vision generation~\cite{melas20243d,zhou2024gala3d,tang2024lgm,pan2024fast,chen2023text,tang2023dreamgaussian,li2023gaussiandiffusion,ren2023dreamgaussian4d} and editing~\cite{zhuang2024tip,chen2023gaussianeditor,fang2023gaussianeditor} tasks. 
Besides, many efforts have been made in anti-aliased rendering~\cite{yu2023mip,yan2023multi} and storage efficiency~\cite{niedermayr2023compressed,fan2023lightgaussian,navaneet2023compact3d,lee2023compact,morgenstern2023compact}.
In this work, further considering the specific imaging characteristics for each virtual camera, we enhance 3DGS~\cite{3DGS} with the ability to specialize the 3D model for a specific camera, thus generating data for the DCSZ task.

\section{Proposed Problem and Method}
In this section, we first introduce the setup of the proposed problem (\ie, dual-camera smooth zoom (DCSZ)) and the motivation of the solution in \cref{section:4_1}.
Then we introduce the way to construct zoom sequences by the proposed data factory in \cref{section:4_2}.
Finally, we provide the details of training a frame interpolation (FI) model for DCSZ with the constructed data in \cref{section:4_3}.

\subsection{Problem Setup and Solution Motivation}
\label{section:4_1}
As illustrated in \cref{fig:intro}(c), the distinct deployment locations and internal characteristics~\cite{zhang2022self,lee2022reference,alzayer2023dc2,abdelhamed2021leveraging} of dual-zoom cameras lead to a notable preview jump in geometric content and image color, which significantly affects the user’s zoom experience.
The proposed dual-camera smooth zoom (DCSZ) aims to achieve a fluid zoom preview, which can be accomplished by a frame interpolation (FI) model to synthesize $N$ intermediate images $\{\tilde{\mathbf{{X}}}_{i}\}_{i=1}^{N}$ between dual-camera images. 
Denote by $\mathbf{X}_{uw}$ and $\mathbf{X}_{w}$ the UW and W image respectively, FI model can be written as, 
\begin{equation}
\tilde{\mathbf{{X}}}_{i} = \mathcal{I}(\mathbf{X}_{uw}, \mathbf{X}_{w}, t_{i};\mathrm{\Theta_{\mathcal{I}}}),
\end{equation}
where $\mathrm{\Theta_{\mathcal{I}}}$ denotes the parameters of the FI model $\mathcal{I}$. 
$t_{i}$ is the temporal encoding that indicates the relative relationship between the synthesized image and the dual-camera images.
Due to the inevitable motion gap between training and dual-camera data, existing pre-trained FI models often fail to meet expectations.
Retraining or fine-tuning the model is a more appropriate way, but it is impractical to capture the ground truth by deploying physical cameras.

To address the issue, we suggest assembling continuous virtual cameras to construct zoom sequences.
For each dual-camera image pair, we expect to construct a set of $N$ virtual cameras $\{v_{i}\}^{N}_{i=1}$ along UW and W cameras.
Thus, virtual ground truth $\{{\mathbf{{X}}}_i\}^{N}_{i=1}$ can be captured by these cameras.
By performing the steps on multiple image pairs, a synthetic dataset for the DCSZ task can be constructed, thus being employed to fine-tune a FI model, \ie,
\begin{equation}
\label{eq:VFI_training_eqn}
\mathrm{\Theta_{\mathcal{I}}^*} = 
\arg \min _{\mathrm{\Theta_{\mathcal{I}}}} \mathcal{L}_{FI}
\left(\mathcal{I}
(\mathbf{X}_{uw}, \mathbf{X}_{w},
t_{i}; 
\mathrm{\Theta_{\mathcal{I}})}, 
{\mathbf{{X}}}_i\right).
\end{equation}
where $\mathcal{L}_{FI}$ denotes the learning objective during training.
The construction of virtual cameras is the key to the solution.
On the one hand, its construction must consider camera extrinsic and intrinsic parameters, as they determine the geometric content.
These parameters can be easily obtained by interpolating the dual-camera corresponding ones.
After that, with the 3D model of the scene, we can generate the captured images.
On the other hand, the construction of the virtual camera must consider the other camera-dependent parameters (\eg, ISP), as they can influence the image color.
Similar to extrinsic and intrinsic parameters, one way that construct these camera-dependent parameters is to interpolate between the dual-camera ones.
However, the general unavailability and complex interpolation of these parameters pose significant challenges.
To circumvent the issues, instead of constructing a single 3D model for all cameras, we propose to construct a specific 3D model for each camera.
In such a case, by inputting the intrinsic and extrinsic parameters of the virtual camera into the camera-specific 3D model, we can obtain the expected image.

Thus, we elaborately design the dual-camera smooth zoom Gaussian Splatting (ZoomGS), which introduces a camera-specific encoding (\ie, $e$ in \cref{fig:pipeline}(b)) as model input to decouple the geometric content of the scene and the camera-dependent characteristics.
Specifically, we take 3DGS~\cite{3DGS} as our 3D representation manner, due to its efficient reconstruction and real-time rendering capabilities.
And we utilize the UW camera 3D  Gaussians as a base 3D model.
As illustrated in \cref{fig:pipeline}(c), when inputting the virtual camera encoding ${e}_{v_{i}}$ (\ie, ${e}={e}_{v_{i}}$), ZoomGS constructs a specific 3D model for the virtual camera. 
Therein, ${e}_{v_{i}}$ can be obtained by the interpolation between ${e}_{uw}$ and ${e}_{w}$.

\begin{figure}[t!]
    \centering
    \includegraphics[width=0.99\linewidth]{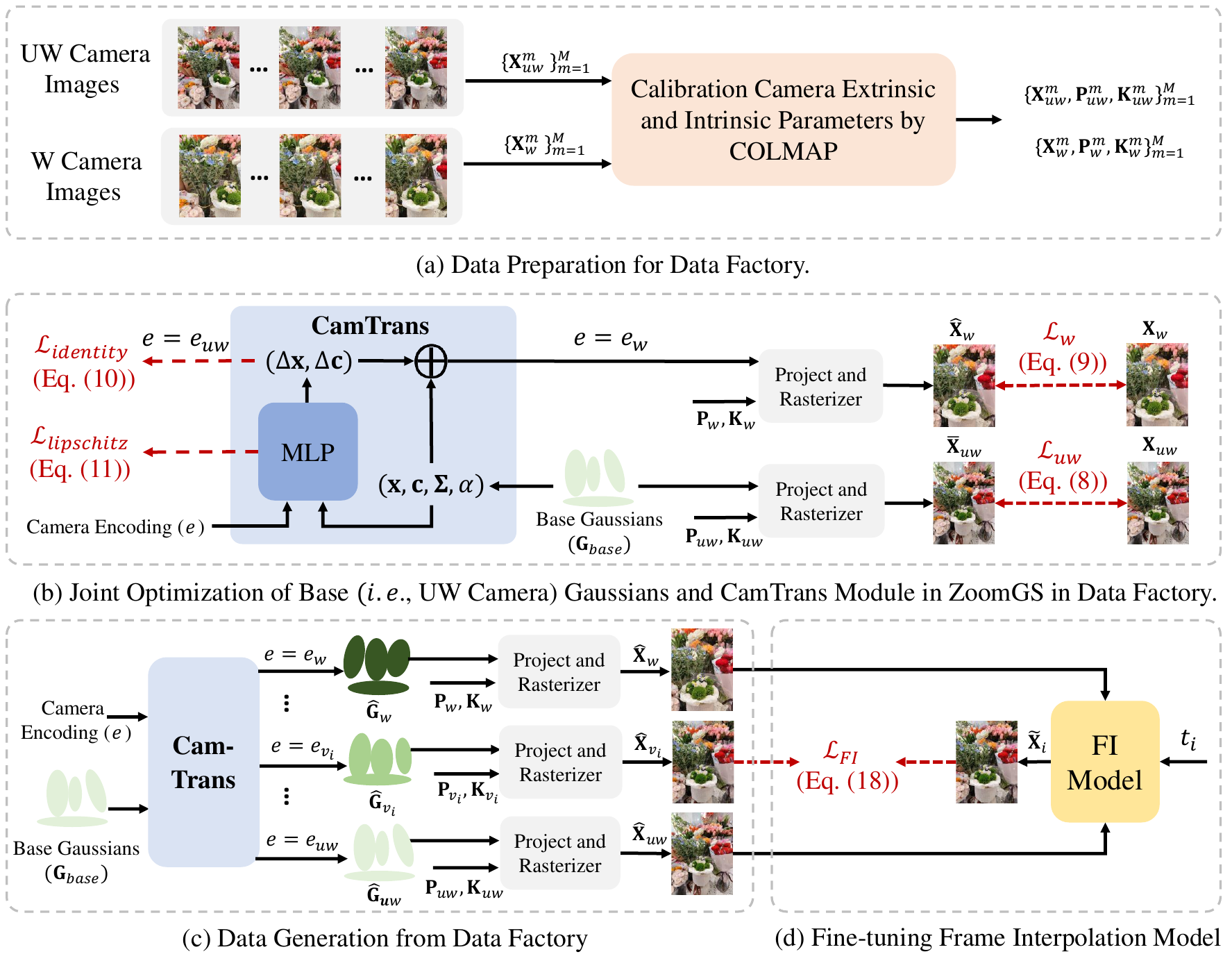}
    \caption{Overview of the proposed method.
    (a) Data preparation for data factory.
    We collect multi-view dual-camera images and calibrate their camera extrinsic and intrinsic parameters.
    (b) Construction of ZoomGS in data factory.
    ZoomGS employs a camera transition (CamTrans) module to transform the base (\ie, UW camera) Gaussians to the specific camera Gaussians according to the camera encoding.
    (c) Data generation from data factory.
    The virtual (V) camera parameters are constructed by interpolating the dual-camera ones, and are then input into ZoomGS to generate zoom sequences.
    (d) Fine-tuning a frame interpolation (FI) model with the constructed zoom sequences.
    }
    \label{fig:pipeline}
\end{figure}

\subsection{Data Factory}
\label{section:4_2}
We briefly introduce the preliminary knowledge for 3D Gaussian Splatting~\cite{3DGS} and then detail the workflow of the proposed data factory. 
In the data factory, we first prepare multi-view dual-zoom images and calibrate their extrinsic and intrinsic parameters (see \cref{fig:pipeline}(a)).
Then we construct ZoomGS models for each scene (see \cref{fig:pipeline}(b)).
After that, we generate 3D models of virtual cameras and render zoom sequences to construct DCSZ data (see \cref{fig:pipeline}(c)).   

\subsubsection{Preliminaries} 
\label{section:4_2_1} 
\ 
\newline
%

\noindent3D Gaussian Splatting (3DGS)~\cite{3DGS} explicitly represents a scene with a set of 3D Gaussians.
Each 3D Gaussian is characterized by its center position $\mathbf{x} \in \mathbb{R}^{3}$ in the world space, shape defined by 3D covariance matrix $\mathbf{\Sigma}$, opacity $\alpha \in \mathbb{R}$, and color defined by anisotropic spherical harmonics (SH) coefficients $\mathbf{c} \in \mathbb{R}^{k}$ ($k$ denotes degrees of SH functions). 
All the characteristics are learnable and optimized through back-propagation.
\noindent\textbf{Project.} 
With the provided camera extrinsic and intrinsic parameters, the 3D Gaussians are projected from the 3D world coordinates into the 2D image coordinates with the covariance matrix $\mathbf{\Sigma}'$, which can be denoted as,
\begin{equation}
\mathbf{\Sigma}^{'} = \mathbf{J}\mathbf{W}\mathbf{\Sigma} \mathbf{W}^{T}\mathbf{J}^{T},
\end{equation}
where $\mathbf{J}$ is the Jacobian of the affine approximation of the projective transformation, and $\mathbf{W}$ is the world-to-camera
matrix~\cite{zwicker2001surface}.
\noindent\textbf{Rasterizer.} After projecting the 3D Gaussians to 2D spaces, the alpha-blending rendering based on point cloud is performed, which is similar to the volumetric rendering in NeRF~\cite{NeRF} in the formulation.
It can be written as,
\begin{equation}
c =\sum_{l \in L} c_l \alpha_l \prod_{j=1}^{l-1}\left(1-\alpha_j\right),
\end{equation}
where $c$ is the pixel value of the rendered image, $L$ is the number of ordered points that overlap the pixel, $c_l$ is the color of a point that is calculated from the SH coefficients $\mathbf{c}$, and $\alpha_l$ is the opacity.
Collectively, 3DGS~\cite{3DGS} can represent a scene with a set of 3D Gaussians, where each one composes of $\{\mathbf{x}, \mathbf{c}, \mathbf{\Sigma}, \alpha\}$.
With the provided camera extrinsic and intrinsic parameters, these Gaussians can be projected into 2D spaces and rasterized to obtain the corresponding rendered image.
We suggest the readers refer to 3DGS~\cite{3DGS} for more detailed descriptions.

\subsubsection{3D Model Construction in Data Factory}
\label{section:4_2_2}
\ 
\newline
%

\noindent\textbf{Data Preparation.}
We follow common procedure in 3D reconstruction works~\cite{3DGS,NeRF, barron2021mip,barron2022mip,muller2022instant} to prepare data.
For each scene, we capture $M$ dual-zoom image pairs from different viewpoints with a mobile phone, where $m$-th pair consists of a UW image $\mathbf{X}^{(m)}_{uw}$ and a W image $\mathbf{X}^{(m)}_{w}$.
Then we feed $\{\mathbf{X}^{(m)}_{uw}\}_{m=1}^{M}$ and $\{\mathbf{X}^{(m)}_{w}\}_{m=1}^{M}$ to COLMAP~\cite{schonberger2016structure} to calibrate the extrinsic and intrinsic parameters for UW images, \ie, $\{\mathbf{P}^{(m)}_{uw}\}_{m=1}^{M}$ and $\{\mathbf{K}^{(m)}_{uw}\}_{m=1}^{M}$, and the two ones for W images,  \ie, $\{\mathbf{P}^{(m)}_{w}\}_{m=1}^{M}$ and $\{\mathbf{K}^{(m)}_{w}\}_{m=1}^{M}$. 
The extrinsic parameters indicate the camera's position and orientation in 3D world coordinates. 
Meanwhile, the intrinsic ones include camera's focal length, influencing the field-of-view (FOV) of the captured images.
In the following sections, for convenience, we denote one pair of dual-camera images as $\mathbf{X}_{uw}$ and $\mathbf{X}_{w}$, the corresponding extrinsic parameters as $\mathbf{P}_{uw}$ and $\mathbf{P}_{w}$, and the corresponding intrinsic ones as $\mathbf{K}_{uw}$ and $\mathbf{K}_{w}$.

\noindent\textbf{Overview of ZoomGS.}
ZoomGS should build multiple camera-specific 3D models.
It is difficult to train each model for each virtual camera independently, as only UW and W images are available.
Instead, ZoomGS transforms the base (\ie, UW camera) 3D model to one for a specific camera according to the camera-specific encoding by camera transition (CamTrans) module.
Denote by $\mathbf{G}_{base}$ the base model, CamTrans can be written as,
\begin{equation}
\label{eq:DZGS_1}
    \mathbf{\hat{G}}_{uw} = \mathcal{T}({e}_{uw}, \mathbf{G}_{base}), \quad
    \mathbf{\hat{G}}_{v_{i}} = \mathcal{T}({e}_{v_{i}}, \mathbf{G}_{base}),\quad
    \mathbf{\hat{G}}_{w} = \mathcal{T}({e}_{w}, \mathbf{G}_{base}).
\end{equation}
$\mathcal{T}$ denotes the CamTrans module. $\mathbf{\hat{G}}_{uw}$, $\mathbf{\hat{G}}_{w}$ and $\mathbf{\hat{G}}_{v_{i}}$ are the 3D models specific to the UW, W and virtual camera $v_{i}$, respectively.
$e_{uw}$, $e_{w}$ and $e_{v_{i}}$ are their camera encodings, respectively.
$e_{uw}$ and $e_{w}$ are set to 0.0 and 1.0, respectively.

Another important issue is to ensure zoom smoothness in ZoomGS.
During training, only $e_{uw}$ and $e_{w}$ are involved.
When using an unseen $e_{v_{i}}$ as input during testing, artifacts easily appear in the rendered images.
Instead, we hope a smooth change of camera encoding leads to a smooth change of 3D models.
In other words, the CamTrans module should be Lipschitz continuous (see suppl. for the details).
To achieve this goal, we introduce a Lipschitz regularization item~\cite{liu2022learning}, which encourages the CamTrans module to be a Lipschitz continuous mapping by penalizing the upper bound on its Lipschitz constant.

With the acquisition of camera-specific 3D models, we take the extrinsic and intrinsic parameters as inputs to render the image for each camera, \ie,
\begin{equation}
\label{eq:DZGS_2}
    \hat{\mathbf{X}}_{uw} \!= \!\mathcal{R}(\mathbf{\hat{G}}_{uw}, \mathbf{P}_{uw}, \mathbf{K}_{uw}),   
    \hat{\mathbf{X}}_{v_{i}}\!= \!\mathcal{R}(\mathbf{\hat{G}}_{v_{i}},\mathbf{P}_{v_{i}}, \mathbf{K}_{v_{i}}),  
    \hat{\mathbf{X}}_{w} \!=\! \mathcal{R}(\mathbf{\hat{G}}_{w},\mathbf{P}_{w}, \mathbf{K}_{w}),
\end{equation}
where $\mathcal{R}$ denotes the rendering process of 3D Gaussians, including projection and rasterization.
$\mathbf{P}_{v_{i}}$ and $\mathbf{E}_{v_{i}}$ are the extrinsic and intrinsic parameters for virtual camera $v_{i}$, respectively.
\noindent\textbf{Camera Transition Module}.
Note that the transformation of 3D models is equivalent to that of 3D Gaussians, as 3D Gaussians explicitly represent the 3D model. 
As shown in \cref{fig:pipeline}(b), given the camera encoding ${e}$ and the properties of base Gaussians $\mathbf{G}_{base}$ (\ie, $\{\mathbf{x}, \mathbf{c}, \mathbf{\Sigma}, \alpha\}$) as inputs, CamTrans module utilizes a tiny MLP to predict the residuals of properties. 
On the one hand, considering that the camera ISP affects image color, we predict color offset $\mathrm{\Delta}\mathbf{c}$.
On the other hand, considering that the distortion degree in images captured by different cameras varies, we also predict the position offset $\mathrm{\Delta}\mathbf{x}$.
Denote by $\mathcal{F}$  the MLP with parameters $\mathrm{\Theta_\mathcal{F}}$, it can be written as,
\begin{equation}
\mathrm{\Delta}\mathbf{x},\mathrm{\Delta}\mathbf{c} = \mathcal{F}(e, \mathbf{x}, \mathbf{c}, \mathbf{\Sigma}, \alpha; \mathrm{\Theta_\mathcal{F}}).
\end{equation}
Thus, the properties of the specific camera Gaussians are $\{\mathbf{x}+\mathrm{\Delta}\mathbf{x}, \mathbf{c}+\mathrm{\Delta}\mathbf{c}, \mathbf{\Sigma}, \alpha\}$.
Please refer to the suppl. for more details about the MLP.
\noindent\textbf{Optimization of ZoomGS.}
The optimization of ZoomGS is illustrated in \cref{fig:pipeline}(b).
The base (\ie, UW camera) Gaussians are initially optimized alone and then jointly optimized with the CamTrans module.
To optimize the base Gaussians, we calculate the reconstruction loss $\mathcal{L}_{uw}$, which can be written as,
\begin{equation}
\label{eq:UW_Reconstrcution}
\mathcal{L}_{uw} = (1 - \beta)\mathcal{L}_{1}(\overline{\mathbf{X}}_{uw}, \mathbf{X}_{uw}) + \beta\mathcal{L}_{ssim}(\overline{\mathbf{X}}_{uw}, \mathbf{X}_{uw}).
\end{equation}
$\overline{\mathbf{X}}_{uw}$ is the rendered image from the base Gaussians.
$\mathcal{L}_{1}$ and $\mathcal{L}_{ssim}$ are $\ell_1 $ loss and SSIM~\cite{wang2004image} loss, respectively.
$\beta$ is set to $0.2$.
The settings all follow 3DGS~\cite{3DGS}.
To optimize the CamTrans module, we employ reconstruction loss for W camera $\mathcal{L}_{w}$, identity regularization item $\mathcal{L}_{identity}$, and Lipschitz regularization item~\cite{liu2022learning} $\mathcal{L}_{lipschitz}$.
First, when ${e}$ is set to ${e}_{w}$, CamTrans module should generate the W camera Gaussians.
It is achieved by  $\mathcal{L}_{w}$, \ie,
\begin{equation}
\label{eq:W_Reconstrcution}
\mathcal{L}_{w} = (1 - \beta)\mathcal{L}_{1}(\hat{\mathbf{X}}_{w}, \mathbf{X}_{w}) + \beta\mathcal{L}_{ssim}(\hat{\mathbf{X}}_{w}, \mathbf{X}_{w}) , \quad if \text{ }{e} = {e}_{w}.
\end{equation}
Second, when ${e}$ is set to ${e}_{uw}$, CamTrans module should generate the UW camera Gaussians.
We employ $\mathcal{L}_{identity}$ to encourage CamTrans module to directly output the base Gaussians, \ie,
\begin{gather}
\label{eq:W_regulation}
\mathcal{L}_{identity} = ||\mathrm{\Delta}\mathbf{x}||_1 + ||\mathrm{\Delta}\mathbf{c}||_1, \quad if \text{ }{e} = {e}_{uw}.
\end{gather}
Third, we hope CamTrans module to be a Lipschitz continuous mapping, thus introducing $\mathcal{L}_{lipschitz}$.
It imposes weight normalization (WN) in each MLP layer of CamTrans module, and then constrains a trainable bound $q_d$ that is positively related to the weight of WN in $d$-th MLP layer, \ie,
\begin{equation}
\label{eq:W_lipschitz}
\mathcal{L}_{lipschitz} = \prod_{d=1}^D  { \texttt{Softplus}}(q_{d}),
\end{equation}
where $D$ is the number of MLP layers, $\texttt{Softplus}$ is the softplus activation function.
Please refer to the suppl. for more details about $\mathcal{L}_{lipschitz}$. 
Overall, combined \cref{eq:UW_Reconstrcution},~\cref{eq:W_Reconstrcution},~\cref{eq:W_regulation} and \cref{eq:W_lipschitz}, the loss terms of optimizing ZoomGS can be written as,
\begin{equation}
\label{eq:loss_DZGS}
\mathcal{L}_{zoom} = \mathcal{L}_{uw} + \mathcal{L}_{w} + \lambda_{identity}\mathcal{L}_{identity} + \lambda_{lipschitz}\mathcal{L}_{lipschitz},
\end{equation}
where $\lambda_{identity}$ and $\lambda_{lipschitz}$ are set to $1.0$ and $1 \times 10^{-7}$ respectively.

\subsubsection{Data Generation from Data Factory}
\label{section:4_2_3}
\ 

%
\noindent After training ZoomGS, we have constructed the camera-specific 3D models.
Next, we construct the parameters of virtual cameras, including extrinsic parameters $\mathbf{P}_{v_{i}}$, intrinsic parameters $\mathbf{K}_{v_{i}}$ and camera encoding $e_{v_{i}}$.
We deploy a total of $N$ virtual cameras, and denote $v_{1}$ and $v_{N}$ as virtual UW and W cameras, respectively.
Following previous works~\cite{liu2021mba,wang2023bad}, for extrinsic parameters $\mathbf{P}_{v_{i}}$, we linearly interpolate $\mathbf{P}_w$ and $\mathbf{P}_{uw}$ in the Lie-algebra of $\mathbf{SE}(3)$, \ie,
\begin{equation}
   \mathbf{P}_{v_{i}} = \mathbf{P}_{uw} \odot \texttt{exp}(\frac{i-1}{N-1} \odot \texttt{log}( \frac{\mathbf{P}_{w}}{\mathbf{P}_{uw}})).
\end{equation}
$i \in \{1,2,...,N\}$ is the relative position index for $i$-th virtual camera $v_{i}$.
$\texttt{exp}$ and $\texttt{log}$ are exponential and logarithmic functions, respectively.
$\odot$ is a pixel-wise multiply operation.
We linearly interpolate $\mathbf{K}_{uw}$ and $\mathbf{K}_{w}$ to get $\mathbf{K}_{v_{i}}$, \ie,
\begin{equation}
    \mathbf{K}_{v_{i}} = (1 - \frac{i-1}{N-1})\odot \mathbf{K}_{uw} + \frac{i-1}{N-1}\odot \mathbf{K}_{w}.
\end{equation}
Similarly, we get the camera encoding ${e}_{v_{i}}$ by the linear interpolation of ${e}_{uw}$ and ${e}_{w}$.
With the constructed parameters of virtual cameras, we can render the zoom sequences between dual cameras from their 3D models according to \cref{eq:DZGS_1,eq:DZGS_2}.
Multiple zoom sequences from diverse scenes will be used to construct a synthetic dataset for fine-tuning FI models.
Please see the \href{https://dualcamerasmoothzoom.github.io}{site} for the visualization of some zoom sequence.

\subsection{Training FI Model for DCSZ Task}
\label{section:4_3}
As shown in \cref{fig:pipeline}(d), with the constructed dataset, we can fine-tune an FI model for DCSZ.
We adopt reconstruction loss $\mathcal{L}_{rec}$ and perceptual loss $\mathcal{L}_{per}$ to optimize the FI model.
Following previous works~\cite{niklaus2018context,niklaus2020softmax,RIFE}, $\mathcal{L}_{rec}$ is calculated between two Laplacian pyramid representations of the reconstructed image $\tilde{\mathbf{X}}_{i}$
and ground truth $\mathbf{X}_{i}$.
Denote by $\texttt{LP}$ the Laplacian decomposition operation, $\mathcal{L}_{rec}$ can be written as,
\begin{equation}
    \mathcal{L}_{rec} = \mathcal{L}_{1}(\texttt{LP}(\tilde{\mathbf{X}}_{i}), \texttt{LP}(\mathbf{X}_{i})).
\end{equation}
The VGG-based perceptual loss is used to improve the visual effect, \ie,
\begin{equation}
    \mathcal{L}_{per} = \mathcal{L}_{1}(\phi(\tilde{\mathbf{X}}_{i}), \phi(\mathbf{X}_{i})),
\end{equation}
where $\phi$ denotes the pre-trained VGG-19~\cite{simonyan2014very} network.
Overall, the loss terms $\mathcal{L}_{FI}$ for fine-tuning FI models can be written as,
\begin{equation}
    \label{eq:VFI_loss}
    \mathcal{L}_{FI} = \mathcal{L}_{rec} + \lambda_{per}\mathcal{L}_{per},
\end{equation}
where $\lambda_{per}$ is set to 1.0 in our experiments.

\section{Experiments}

\subsection{Experimental Settings}

\noindent\textbf{Datasets.}
The multi-view dual-camera image pairs for reconstructing 3D models are captured by Redmi K50 Ultra mobile phone, including 48 indoor ones captured in \textit{classrooms}, \textit{dining halls}, and \textit{shopping malls}, and 30 outdoor ones collected in \textit{campuses} and \textit{companies}.
For each scene, we first capture dual-camera image pairs from 15 front-facing viewpoints.
Due to the narrower FOV of W camera, surrounding areas in the UW image are not seen in the corresponding W image. 
To cover these areas, we additionally capture 6 images with W camera.
In each scene, we randomly select 5 dual-camera image pairs to evaluate the performance of ZoomGS, and the remaining images are used for training.
Based on the proposed data factory, we generate a synthetic dataset for DCSZ, which comprises $155$ zoom sequences.
127 sequences from 64 randomly sampled scenes are used for training, and the remaining 28 sequences from 14 scenes are used for testing.
Each sequence contains $33$ frames with a resolution $1632 \times 1216$.
In addition, we collect a real-world dataset from 100 scenes that are non-overlapped with the synthetic dataset to evaluate FI model in the real world.
Please see some dataset examples in the Suppl.

\begin{table*}[t!] 
    \small
    \renewcommand\arraystretch{1}
    \begin{center}
    \caption{
        Quantitative comparisons of FI models on the synthetic dataset and real-world dataset.  
        $\uparrow$ denotes the higher metric the better, and $\downarrow$ denotes the lower one the better.
        The results of our fine-tuned models are marked \pmb{bold}.
    } 
    \label{tab:sota_comparison}
    \scalebox{0.8}{
    \begin{tabular}{p{2.5cm}<{\centering} p{5cm}<{\centering} p{5cm}<{\centering}} 
    \toprule
      \multirow{2}{*}{Methods} & Synthetic & Real-World\\
     & {PSNR}{$\uparrow$}/{SSIM}{$\uparrow$}/{LPIPS}{$\downarrow$} & NIQE$\downarrow$/PI$\downarrow$/CLIP-IQA$\uparrow$/MUSIQ$\uparrow$\\
        \midrule
        EDSC~\cite{EDSC} & 20.91 / 0.7125 / 0.258 & 4.40 / 4.93 / 0.3775 / 49.917  \\
        Ours (EDSC)  & \pmb{22.28} / \pmb{0.7430} / \pmb{0.114}& \pmb{4.05} / \pmb{3.93} / \pmb{0.5592} / \pmb{64.873} \\
        \midrule
        IFRNet~\cite{IFRNet} & 20.27 / 0.6919 / 0.233 & 4.25 / 4.55 / 0.4154 / 54.724 \\  
        Ours (IFRNet)  & \pmb{22.54} / \pmb{0.7570} / \pmb{0.109} & \pmb{3.99} / \pmb{3.90} / \pmb{0.5231} / \pmb{63.792} \\
        \midrule
        RIFE~\cite{RIFE} &  22.03 / 0.7426 / 0.234 & 4.94 / 5.43 / 0.3560 / 50.168   \\  
        Ours (RIFE)  & \pmb{22.95} / \pmb{0.7661} / \pmb{0.097}  & \pmb{3.84} / \pmb{3.74} / \pmb{0.5605} / \pmb{65.383}\\
        \midrule
        AMT~\cite{AMT} &  20.80 / 0.7038 / 0.218 & 5.11 / 5.32 / 0.3966 / 55.151\\
        Ours (AMT) & \pmb{23.05} / \pmb{0.7746} / \pmb{0.098} & \pmb{4.00} / \pmb{3.85} / \pmb{0.5358} / \pmb{65.268}\\
        \midrule
        UPRNet~\cite{UPRNet} & 19.46 / 0.7029 / 0.224  & 3.99 / 4.38 / 0.4142 / 57.959\\
        Ours (UPRNet)  & \pmb{23.61} / \pmb{0.7808} / \pmb{0.101} & \pmb{3.98} / \pmb{3.84} / \pmb{0.5537} / \pmb{63.490}\\
        \midrule
        EMAVFI~\cite{EMAVFI}  &  22.21 / 0.7527 / 0.220 & 5.00 / 5.45 / 0.3788 / 51.864\\
        Ours (EMAVFI) & \pmb{23.76} / \pmb{0.7912} / \pmb{0.088} & \pmb{3.64} / \pmb{3.65} / \pmb{0.5413} / \pmb{63.906}\\ 
	\bottomrule
	\end{tabular}
        }
    \end{center}
\end{table*}

\noindent\textbf{Training Configurations.}
We adopt Adam optimizer~\cite{kingma2014adam} to train the base Gaussians for 5k iterations first and then jointly optimize it with CamTrans module for 30k iterations.
The learning rate setting for base Gaussians is the same as 3DGS~\cite{3DGS}, while the learning rate for CamTrans module is initially set to $1\times10^{-3}$ and decayed to $1\times10^{-6}$ after 20k iterations.

For FI models, we fine-tune the pre-trained models for 50k iterations.
We augment the training data with random horizontal and vertical flips.
Cosine annealing strategy~\cite{loshchilov2016sgdr} is employed to decrease the learning rate from $1.5\times10^{-4}$ to $1.5\times10^{-6}$ steadily.
All experiments are conducted with PyTorch~\cite{paszke2019pytorch} on two Nvidia GeForce RTX A6000 GPUs.
\noindent\textbf{Evaluation Configurations.}
For ZoomGS, we calculate PSNR~\cite{huynh2008scope}, SSIM~\cite{wang2004image} and LPIPS~\cite{zhang2018unreasonable} metrics between rendered images and dual-camera images.
For FI models, we employ PSNR~\cite{huynh2008scope}, SSIM~\cite{wang2004image} and LPIPS~\cite{zhang2018unreasonable} metrics to evaluate the performance on the synthetic dataset.
Because there is no ground truth in the real-world data, we employ multiple no-reference metrics for evaluation, \ie, NIQE~\cite{mittal2012making}, PI~\cite{blau20182018}, CLIP-IQA~\cite{wang2023exploring}, and MUSIQ~\cite{ke2021musiq}.

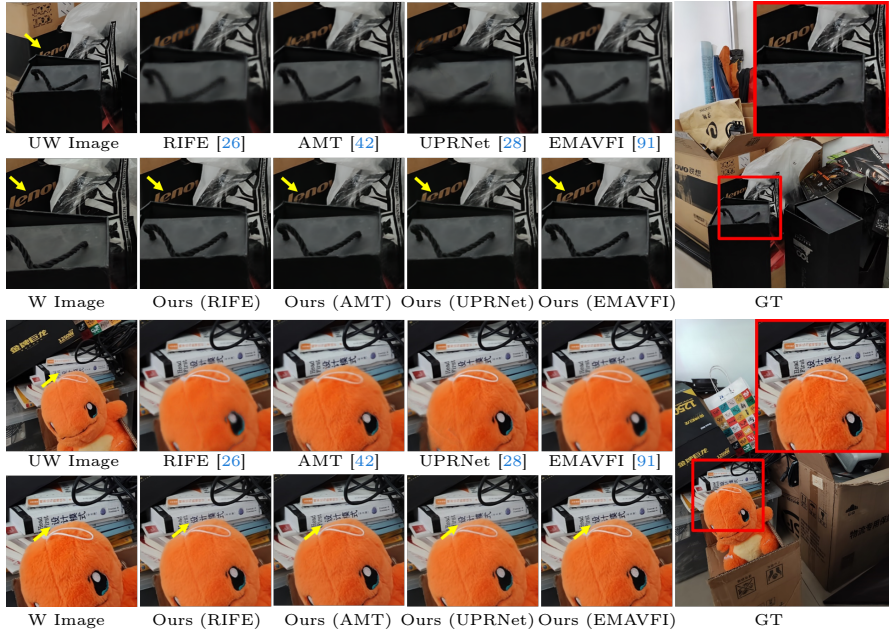
\begin{figure}[t!]
    \centering
    \begin{subfigure}{0.98\textwidth}
        \ContinuedFloat
        \begin{overpic}[width=0.98\textwidth,grid=False]
        {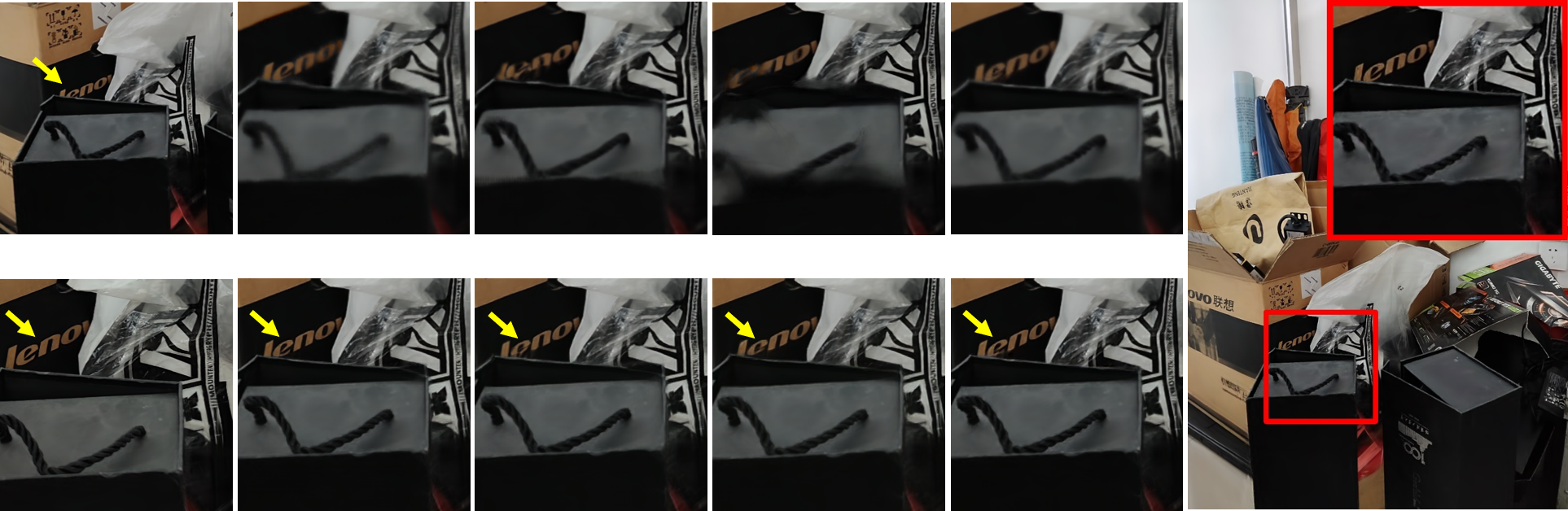}
         \put(8,53){{\fontsize{6.5}{7}\selectfont UW Image}}
         \put(59,53){{\fontsize{6.5}{7}\selectfont RIFE~\cite{RIFE}}}
         \put(110,53){{\fontsize{6.5}{7}\selectfont AMT~\cite{AMT}}}
         \put(156,53){{\fontsize{6.5}{7}\selectfont UPRNet~\cite{UPRNet}}}
         \put(205,53){{\fontsize{6.5}{7}\selectfont EMAVFI~\cite{EMAVFI}}}

         \put(8,-7){{\fontsize{6.5}{7}\selectfont W Image}}
         \put(55,-7){{\fontsize{6.5}{7}\selectfont Ours (RIFE)}}
         \put(105,-7){{\fontsize{6.5}{7}\selectfont Ours (AMT)}}
         \put(149,-7){{\fontsize{6.5}{7}\selectfont Ours (UPRNet)}}
         \put(201,-7){{\fontsize{6.5}{7}\selectfont Ours (EMAVFI)}}
          \put(283,-7){{\fontsize{6.5}{7}\selectfont GT}}
        \end{overpic}
    \end{subfigure}

    \caption{Visual comparisons on the synthetic dataset.
    The FI models synthesize the intermediate geometry content between dual cameras, as indicated with yellow arrows. 
    The fine-tuned FI models generate more photo-realistic details.
    } 
    \label{fig:syn_vis_comp}
\end{figure}

\begin{figure}[t!]
    \centering
    \begin{subfigure}{0.99\textwidth}
        \ContinuedFloat
        \begin{overpic}[width=0.99\textwidth,grid=False]
        {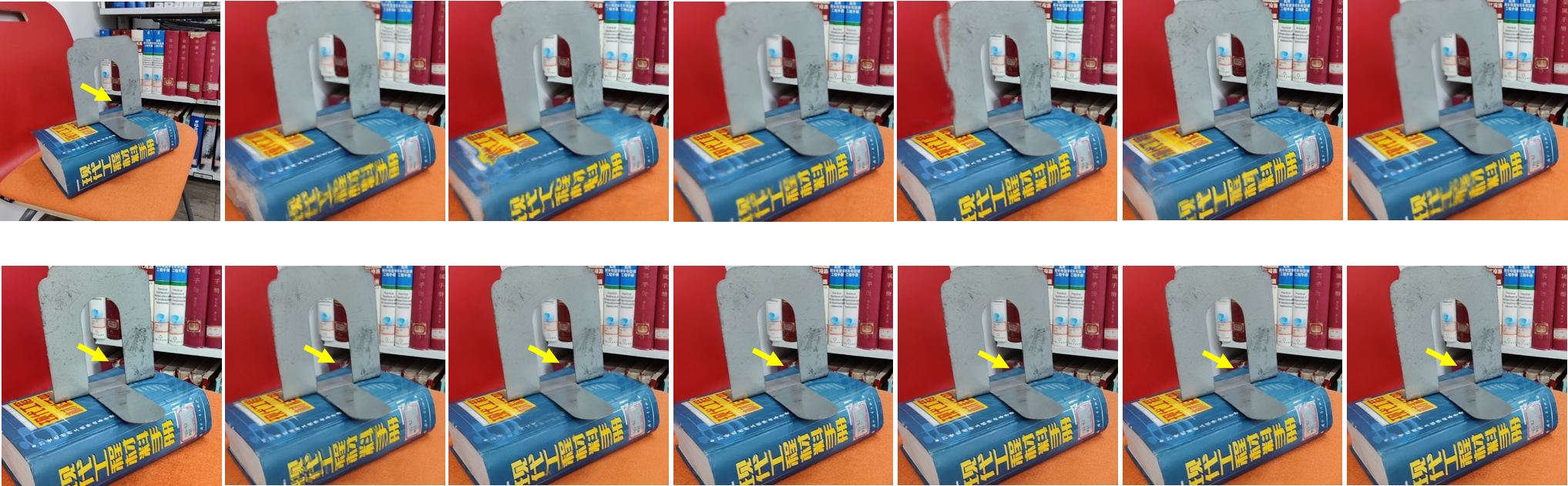}
        \put(7,51){{\fontsize{6.5}{7}\selectfont UW Image}}
        \put(55,51){{\fontsize{6.5}{7}\selectfont EDSC~\cite{EDSC}}}
        \put(100,51){{\fontsize{6.5}{7}\selectfont IFRNet~\cite{IFRNet}}}
        \put(155,51){{\fontsize{6.5}{7}\selectfont RIFE~\cite{RIFE}}}
        \put(204,51){{\fontsize{6.5}{7}\selectfont AMT~\cite{AMT}}}
        \put(247,51){{\fontsize{6.5}{7}\selectfont UPRNet~\cite{UPRNet}}}
        \put(295,51){{\fontsize{6.5}{7}\selectfont EMAVFI~\cite{EMAVFI}}}
        \put(9,-7){{\fontsize{6.5}{7}\selectfont W Image}}
        \put(50,-7){{\fontsize{6.5}{7}\selectfont Ours (EDSC)}}
        \put(98,-7){{\fontsize{6.5}{7}\selectfont Ours (IFRNet)}}
        \put(149,-7){{\fontsize{6.5}{7}\selectfont Ours (RIFE)}}
        \put(195,-7){{\fontsize{6.5}{7}\selectfont Ours (AMT)}}
        \put(238,-7){{\fontsize{6.5}{7}\selectfont Ours (UPRNet)}}
        \put(289,-7){{\fontsize{6.5}{7}\selectfont Ours (EMAVFI)}}
        \end{overpic}
    \end{subfigure}

    \caption{Visual comparisons on the real-world dataset.
    The FI models also synthesize the intermediate geometry content in the real world, as indicated with yellow arrows. 
    Besides, the fine-tuned FI models generate fewer visual artifacts.
    } 
    \label{fig:real_vis_comp}
    \vspace{-6mm}
\end{figure}

\subsection{Comparison with State-of-the-Art Methods}
\noindent\textbf{Comparison Configurations}.
We conduct experiments with 6 state-of-the-art FI methods (\ie, EDSC~\cite{EDSC}, IFRNet~\cite{IFRNet}, RIFE~\cite{RIFE}, AMT~\cite{AMT}, UPRNet~\cite{UPRNet}, and EMAVFI~\cite{EMAVFI}) to validate the effectiveness of the proposed data factory.
Their pre-trained models are trained on commonly used Vimeo90K~\cite{xue2019video} dataset.
For each method, we fine-tune the pre-trained model as \cref{eq:VFI_loss} on the constructed synthetic dataset, obtaining our models.
\noindent\textbf{Results on Synthetic Dataset}.
We summarize the quantitative results in \cref{tab:sota_comparison}.
After fine-tuning, all methods demonstrate significant performance improvements for DCSZ compared to the original models, which confirms the effectiveness of the proposed data factory.
Taking EMAVFI~\cite{EMAVFI} as an example, it obtains 1.55dB PSNR gain.
Moreover, the qualitative results in \cref{fig:syn_vis_comp} show that the fine-tuned models predict more photo-realistic details.
More visual comparisons can be seen in the suppl.

\noindent\textbf{Results on Real-World Dataset}.
We also evaluate the models' performance in real-world scenarios.
as shown in \cref{tab:sota_comparison}, the fine-tuned models generalize to the real world well and still outperform the pre-trained models by a large margin.
In addition, the fine-tuned models generate fewer visual artifacts as shown in \cref{fig:real_vis_comp}.
It indicates that the synthetic data generated by our data factory is close to the real-world ones.
More visual comparisons can be seen in the suppl.

\section{Ablation Study}
The ablation experiments are mainly conducted to validate the effectiveness of ZoomGS, which is the key to our data factory.
The experiments are performed with 10 randomly sampled scenes from the prepared multi-view dual-camera images.
We first compare ZoomGS with 3DGS~\cite{3DGS}.
Then, we validate the effect of identity and Lipschitz regularization terms, as well as the MLP in ZoomGS.

\begin{table*}[t!] 
    \small
    \renewcommand\arraystretch{1}
    \begin{center}
	\caption{
         Comparisons between ZoomGS and 3DGS~\cite{3DGS}.
         We calculate evaluation metrics for UW and W cameras respectively.
         `Average' indicates the averaged results.
         } 
	\label{tab:GZGS_3DGS}
        \scalebox{0.85}{
	\begin{tabular}{ccccccc}
		\toprule
        Methods & \tabincell{c}{Training Data \\ UW / W }
   & \tabincell{c}{PSNR$\uparrow$/SSIM$\uparrow$/LPIPS$\downarrow$ \\ UW} & \tabincell{c}{PSNR$\uparrow$/SSIM$\uparrow$/LPIPS$\downarrow$ \\ W} & \tabincell{c}{PSNR$\uparrow$/SSIM$\uparrow$/LPIPS$\downarrow$ \\ Average}\\
        \midrule
        \multirow{3}{*}{\begin{tabular}[c]{@{}c@{}} 3DGS   \\ \end{tabular}} & $\times$ / $\checkmark$ & 17.58/0.5701/0.334 & 
        26.09/0.8433/\pmb{0.183} &
        21.84/0.7067/0.259
        \\
          & $\checkmark$ / $\times$ &  28.78/0.8961/\pmb{0.157}& 
            18.56/0.5638/0.350&
            23.67/0.7300/0.254
            \\
        & $\checkmark$ / $\checkmark$ &  26.03/0.8520/0.194 &  
        25.89/0.8343/0.199 &
        25.96/0.8432/0.197
        \\
        \midrule
        ZoomGS & $\checkmark$ / $\checkmark$ & \pmb{28.79}/\pmb{0.8911}/0.167 & 
        \pmb{26.59}/\pmb{0.8450}/0.198 &
        \pmb{27.69}/\pmb{0.8681}/\pmb{0.183}
        \\
		\bottomrule
	\end{tabular}
        }
    \end{center}
    \vspace{-4mm}
\end{table*}

\begin{table*}[t!] 
    \small
    \renewcommand\arraystretch{1}
    \begin{center}
	\caption{
        Ablation studies on identity regularization term $\mathcal{L}_{identity}$ (see \cref{eq:W_regulation}) and Lipschitz regularization term $\mathcal{L}_{lipschitz}$(see \cref{eq:W_lipschitz}).
        } 
	\label{tab:loss_term}
        \vspace{-2mm}
        \scalebox{0.85}{
	\begin{tabular}{ccccccc}
		\toprule		$\mathcal{L}_{identity}$ & $\mathcal{L}_{lipschitz}$ &  \tabincell{c}{PSNR$\uparrow$/SSIM$\uparrow$/LPIPS$\downarrow$ \\ UW} &
      \tabincell{c}{PSNR$\uparrow$/SSIM$\uparrow$/LPIPS$\downarrow$ \\ W} & \tabincell{c}{PSNR$\uparrow$/SSIM$\uparrow$/LPIPS$\downarrow$ \\ Average}\\
        \midrule
        $\times$ & $\times$& 20.28/0.6739/0.283 & \pmb{26.72}/0.8465/\pmb{0.197} &
        23.50/0.7602/0.240&
        \\
        $\times$ & $\checkmark$& 20.35/0.6703/0.279 & 26.69/\pmb{0.8466}/\pmb{0.197} &
        23.52/0.7585/0.238 &
        \\
        $\checkmark$ & $\times$ & 28.65/0.8896/0.168 & 26.60/0.8448/0.198 & 
        27.63/0.8672/0.183
        \\
        \midrule
        $\checkmark$ & $\checkmark$ & \pmb{28.79}/\pmb{0.8911}/\pmb{0.167} & 26.59/0.8450/0.198 &
        \pmb{27.69}/\pmb{0.8681}/\pmb{0.183}\\
		\bottomrule
	\end{tabular}
        }
    \end{center}
    \vspace{-4mm}
\end{table*}

\subsection{Comparison ZoomGS with 3DGS~\cite{3DGS}}
\cref{tab:GZGS_3DGS} shows the results.
First, when applying a 3DGS model trained with one camera data to the other camera, the performance significantly drops, due to their different imaging characteristics.
Secondly, the effectiveness is still limited when mixing dual-camera data to train a 3DGS model. 
Third, by constructing 3D models for each camera, ZoomGS has better performance. 

\subsection{Effect of Identity and Lipschitz Regularization}
\label{section:regulation}
\begin{figure}[t!]
    \centering
    \includegraphics[width=0.99\linewidth]{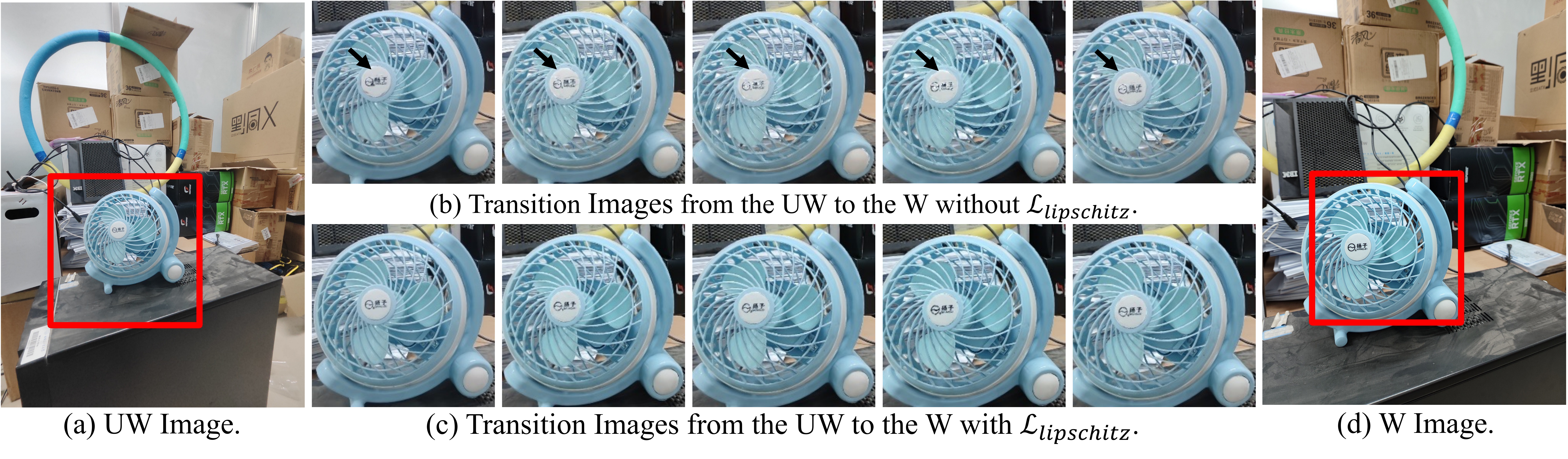}
    \caption{
    Effect of Lipschitz regularization for ZoomGS. 
    Without $\mathcal{L}_{lipschitz}$, it easily produces some visual artifacts, as indicated with black arrows in (b).
    }
    \label{fig:lip_ab}
\end{figure}

Without $\mathcal{L}_{identity}$ (see \cref{eq:W_regulation}), ZoomGS fails to output a 3D model specific to UW camera when camera encoding $e$ is set to ${e}_{uw}$, as shown in \cref{tab:loss_term}.
Without $\mathcal{L}_{lipschitz}$(see \cref{eq:W_lipschitz}), some visual artifacts are produced in virtual camera rendering images, as shown in \cref{fig:lip_ab}(b). 
$\mathcal{L}_{lipschitz}$ significantly improves zoom smoothness by encouraging CamTrans module to be a Lipschitz continuous mapping, as shown in \cref{fig:lip_ab}(c).

\subsection{Effect of MLP}
Here we conduct experiments to validate the effect of predicting different properties in the MLP of CamTrans module.
As shown in \cref{tab:CCTM_output},  predicting residuals of fewer properties (\ie, only $\mathrm{\Delta}\mathbf{c}$ and only $\mathrm{\Delta}\mathbf{x}$) or more properties (\ie, $\{\mathrm{\Delta}\mathbf{c},\mathrm{\Delta}\mathbf{x},\mathrm{\Delta}\mathbf{\Sigma}\}$,
$\{\mathrm{\Delta}\mathbf{c},\mathrm{\Delta}\mathbf{x},\mathrm{\Delta}\alpha\}$,
and $\{\mathrm{\Delta}\mathbf{c},\mathrm{\Delta}\mathbf{x},\mathrm{\Delta}\mathbf{\Sigma},\mathrm{\Delta}\alpha\}$) leads to a slight performance drop.
For simplicity and explainability, CamTrans module predicts ${\rm \Delta}\mathbf{x}$ and ${\rm \Delta}\mathbf{c}$ as 3D model residuals.

\begin{table*}[t!] 
    \small
    \renewcommand\arraystretch{1}
    \begin{center}
	\caption{Ablation studies on the MLP outputs in CamTrans module.
     It predicts the offsets between the 3D model specific to the virtual camera and the base model.
 } 
	\label{tab:CCTM_output}
        \vspace{-2mm}
        \scalebox{0.90}{
	\begin{tabular}{lcccccc}
		\toprule
	  Outputs &  \tabincell{c}{PSNR$\uparrow$/SSIM$\uparrow$/LPIPS$\downarrow$ \\ UW} & \tabincell{c}{PSNR$\uparrow$/SSIM$\uparrow$/LPIPS$\downarrow$ \\ W} & \tabincell{c}{PSNR$\uparrow$/SSIM$\uparrow$/LPIPS$\downarrow$ \\ Average}\\
        \midrule
        $\mathrm{\Delta}\mathbf{c}$  & \pmb{28.85}/\pmb{0.8927}/0.167 & 26.36/0.8358/0.215 &
        27.60/0.8643/0.191
        \\
        $\mathrm{\Delta}\mathbf{x}$  & 28.49/0.8883/0.170 & 26.46/0.8443/0.198 &
        27.48/0.8663/0.184
        \\
        $\mathrm{\Delta}\mathbf{c}$,$\mathrm{\Delta}\mathbf{x}$,$\mathrm{\Delta}\mathbf{\Sigma}$ &  
        28.60/0.8905/0.167& 26.44/0.8444/0.198&
        27.52/0.8675/0.183
        \\
    $\mathrm{\Delta}\mathbf{c}$,$\mathrm{\Delta}\mathbf{x}$,$\mathrm{\Delta}\alpha$ &  28.70/0.8905/0.167 & \pmb{26.60}/\pmb{0.8453}/\pmb{0.197} &
            27.65/0.8679/0.183
        \\

$\mathrm{\Delta}\mathbf{c}$,$\mathrm{\Delta}\mathbf{x}$,$\mathrm{\Delta}\mathbf{\Sigma}$,$\mathrm{\Delta}\alpha$ &  
        28.72/0.8918/\pmb{0.165}& 26.50/0.8444/\pmb{0.197}&
        27.61/\pmb{0.8681}/\pmb{0.182}
        \\
        
        \midrule
        $\mathbf{\mathrm{\Delta} c}$,$\mathbf{\mathrm{\Delta} x}$ & 
        28.79/0.8911/0.167 & 26.59/0.8450/0.198 & \pmb{27.69}/\pmb{0.8681}/0.183
        \\
		\bottomrule
	\end{tabular}
 }
    \end{center}
    \vspace{-6mm}
\end{table*}

\section{Conclusion}
In this work, we introduce a new task, \ie, dual-camera smooth zoom (DCSZ) to achieve a fluid zoom preview on mobile phones.
The frame interpolation (FI) technique is a potential solution but struggles with the ground truth collection using physical cameras.
To address the issue, we suggest a data factory solution that assembles continuous virtual cameras to construct DCSZ data.
In particular, we propose a novel dual-camera smooth zoom Gaussian Splatting (ZoomGS), where each virtual camera has its 3D model. The zoom sequences can be rendered from these 3D models according to the interpolated parameters between dual camera ones.
Moreover, we generate a synthetic dataset for DCSZ based on the data factory for fine-tuning FI models, and capture real-world dual-zoom images for evaluation.
Experiments show that the proposed method significantly improves the performance of FI models on DCSZ task.

\section*{Acknowledgement}
This work was supported in part by the National Natural Science Foundation of China (NSFC) under Grant No. 62371164 and No. U22B2035.

\clearpage


\title{Dual-Camera Smooth Zoom on Mobile Phones \\  (Supplementary Material)}   

\author{Renlong Wu 
\and
Zhilu Zhang$^{(}$\Envelope$^)$ 
\and
Yu Yang
\and
Wangmeng Zuo 
}

\authorrunning{R.Wu et al.}

\institute{
Harbin Institute of Technology, China
\\
\email{hirenlongwu@gmail.com,cszlzhang@outlook.com,\\806224005qq@gmail.com,wmzuo@hit.edu.cn}
}
\maketitle

\renewcommand{\thesection}{\Alph{section}}
\renewcommand{\thetable}{\Alph{table}}
\renewcommand{\thefigure}{\Alph{figure}}
\renewcommand{\thealgorithm}{\Alph{algorithm}}
\renewcommand{\theequation}{\Alph{equation}}

\noindent{The content of the supplementary material involves:}
\begin{itemize}
\item More details of camera transition module in Sec.~\ref{sec:supp_1}.
\item More details of Lipschitz regularization item in Sec.~\ref{sec:supp_2}.
\item More details of synthetic and real-world datasets in Sec.~\ref{sec:supp_datasets}.
\item More visual results of ZoomGS in Sec.~\ref{sec:supp_zoomgs}.
\item More visual results of FI models in Sec.~\ref{sec:supp_3}.
\item Limitation in Sec.~\ref{sec:supp_4}.
\end{itemize}

\section{More Details of Camera Transition Module}
\label{sec:supp_1}

\begin{figure}[t!]
    \centering
    \includegraphics[width=0.6\linewidth]{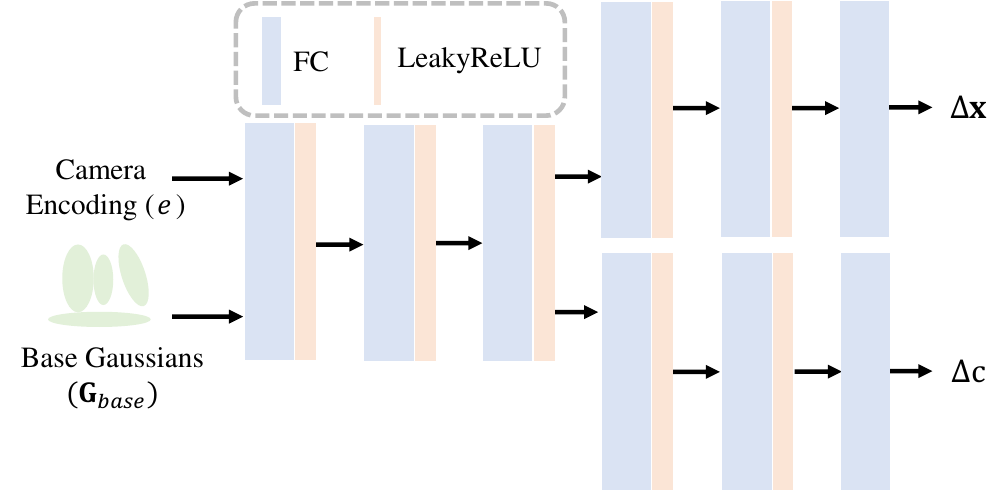}
    \caption{Illustration of MLP in CamTrans module.}
    \label{fig:mlp}
\end{figure}

The architecture of MLP in Camera Transition (CamTrans) module is provided in~\cref{fig:mlp}.
It stacks three FC blocks as the main branch,  where each block consists of an FC layer followed by an LeakyReLU operation.
Then we deploy two heads to predict position offsets $\mathbf{\Delta x}$ and color offsets  $\mathbf{\Delta c}$, respectively.

\section{More Details of Lipschitz Regularization Item}
\label{sec:supp_2}
\noindent\textbf{Lipschitz Continuous}.
A neural network $f_{\theta}$ with parameter $\theta$ is called Lipschitz continuous if there exist a constant $q \geq 0$ such that
\begin{align}
  \underbrace{\| f_\theta(e_0) - f_\theta(e_1) \|_p}_{\mathclap{\text{change in the output}}} \leq q\ \underbrace{\| e_0 - e_1\|_p}_{\mathclap{\text{change in the input}}}
\end{align}
for all possible inputs $e_0$ and $e_1$ under a $p$-norm choice.
The parameter $q$ is called the Lipschitz constant.
In the CamTrans module, we hope a smooth change of camera encoding leads to a smooth change of 3D models, thus, we introduce the Lipschitz regularization item~\cite{liu2022learning} to encourage it to be a Lipschitz continuous mapping. 

\noindent\textbf{Lipschitz Regularization Item}.
Denote $\mathcal{L}_{lipschitz}$ by the Lipschitz regularization item.
Following previous work~\cite{liu2022learning}, we impose an Lipschitz weight normalization (WN) in each MLP layer of CamTrans module, as shown in \cref{alg:WM}.
Then, we introduce $\mathcal{L}_{lipschitz}$ on per-layer Lipschitz bounds in WN, \ie,
\begin{equation}
\label{eq:W_lipschitz_supp}
\mathcal{L}_{lipschitz} = \prod_{d=1}^D  { \texttt{Softplus}}(q_{d}).
\end{equation}
$q_{d}$ is a trainable bound for $d$-th WN layer,
$\texttt{Softplus}$ is the softplus activation function, 
$\texttt{Softplus}(q_{d})$ is the Lipschitz bound for  $d$-th WN layer.
$D$ is the number of MLP layers.

\begin{algorithm*}[t!]
\caption{Pseudo code about Lipschitz weight normalization (WN) }
\begin{algorithmic}[1]    
    \Require $\mathbf{W}_{d}$: weight of $d$-th MLP layer, \\
    $q_{d}$: trainable bound.
    \State
    $\mathbf{\hat{w}}_{d} = \sum_{c=1}^{C_{in}}|\mathbf{W}_{d}^{c}|$ \Comment{$C_{in}$: input channel number of $d$-th MLP layer}
    \State
    $\mathbf{s} = \texttt{min}(\texttt{Softplus}(q_d) \odot \frac{1}{\mathbf{\hat{w}}_{d}}, 1.0)$
    \State
    \Return $\mathbf{s} \odot \mathbf{W}_{d} $
 
\end{algorithmic}
\label{alg:WM}
\end{algorithm*}

\section{More Details of Synthetic and Real-World Datasets}
\label{sec:supp_datasets}
\begin{figure}[t!]
    \centering
    \includegraphics[width=0.98\linewidth]{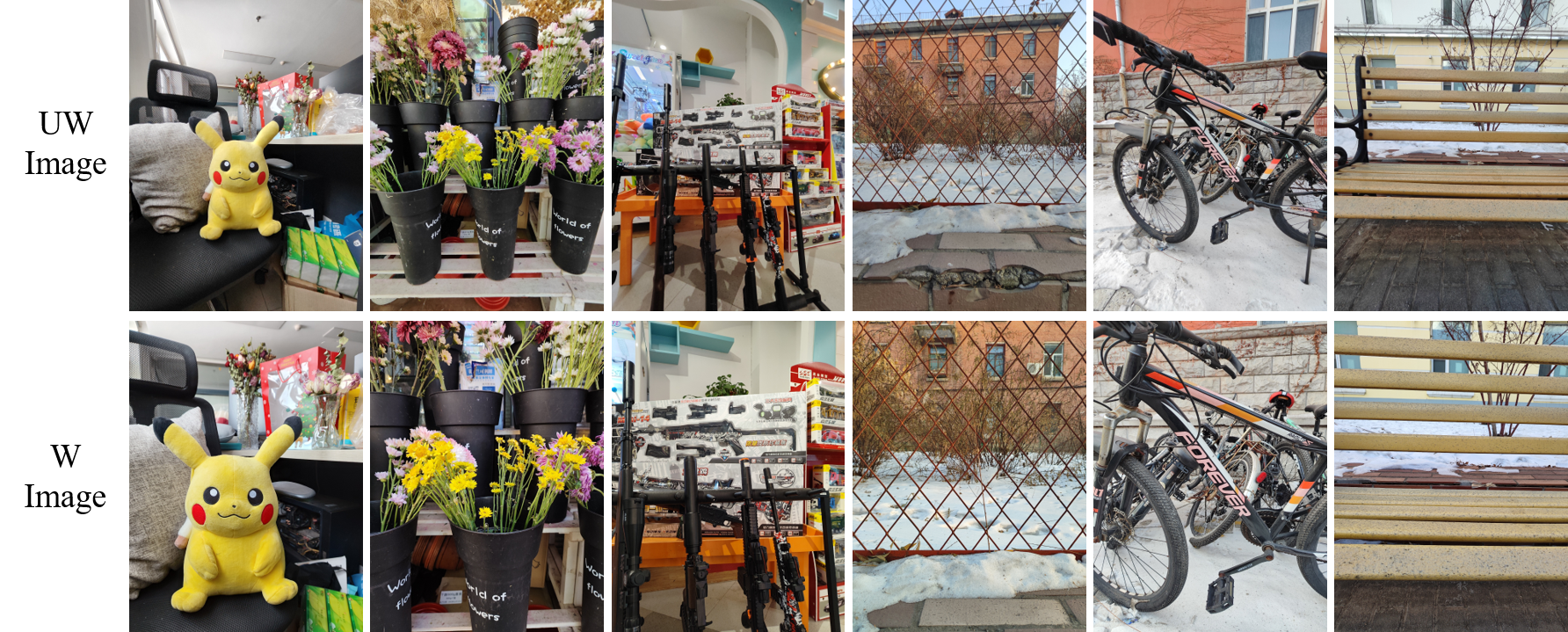}
    \caption{Some examples from synthetic datasets.}
    \label{fig:synthetic_dataset_example}
\end{figure}

\noindent\textbf{Synthetic dataset.} We generate 155 synthetic zoom sequences for DCSZ from 78 scenes, including 48 indoor ones captured in \textit{classrooms}, \textit{dining halls}, and \textit{shopping malls}, and 30 outdoor ones collected in \textit{campuses} and \textit{companies}.
Some examples of the synthetic dataset are shown in \cref{fig:synthetic_dataset_example}.
127 sequences from 64 randomly sampled scenes are used for training, and the remaining 28 sequences from 14 scenes are used for testing.
\noindent\textbf{Real-World dataset.} We additionally capture dual camera images from 100 scenes that are non-overlapped with the synthetic dataset to evaluate FI model in the real world.
It includes diverse scenes, like \textit{desks}, \textit{chairs}, \textit{debris piles}, \textit{billboards}, \textit{buildings}, \textit{vegetation}, \etc.
Some examples of the real-world dataset are shown in \cref{fig:real_dataset_example}.

\begin{figure}[t!]
    \centering
    \includegraphics[width=0.98\linewidth]{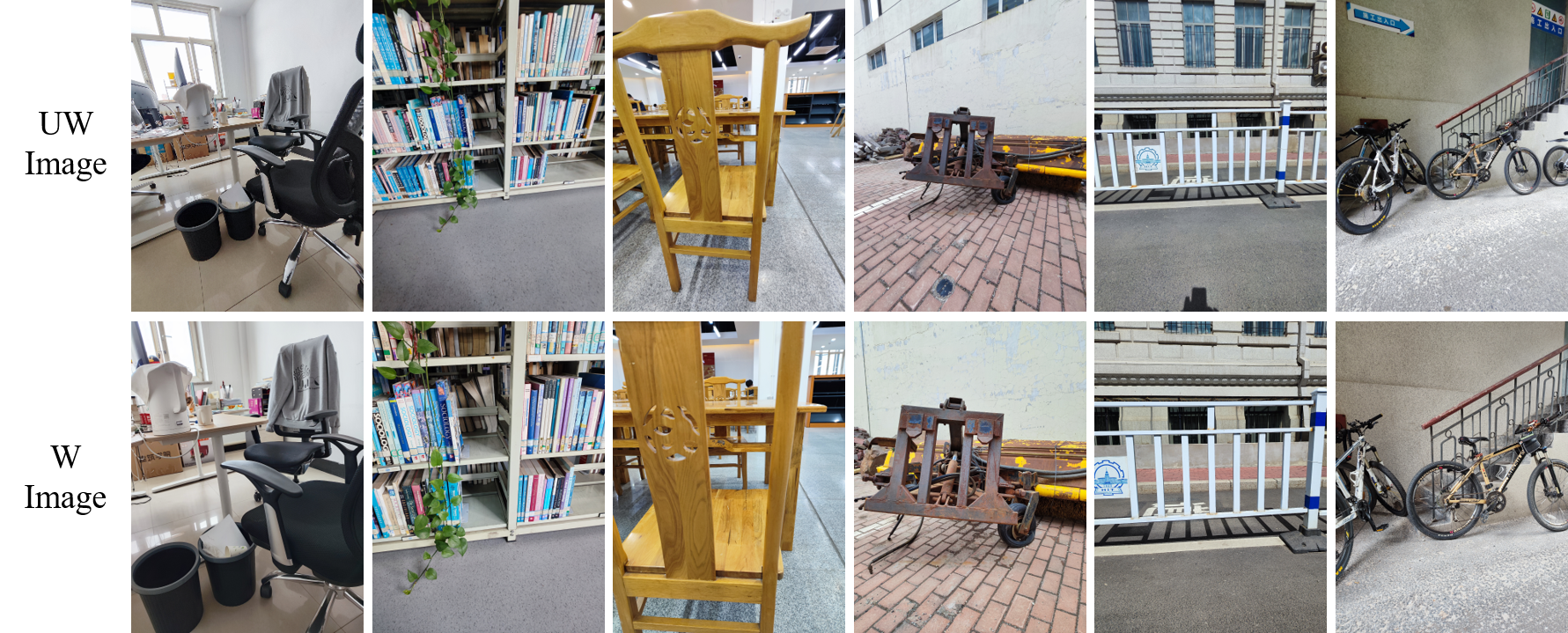}
    \caption{Some examples from real-world datasets.}
    \label{fig:real_dataset_example}
\end{figure}

\section{More Visual Results of ZoomGS}
\label{sec:supp_zoomgs}
We provide the visual comparison results between 3DGS~\cite{3DGS} and the proposed ZoomGS in \cref{fig:more_vis_comp_zoomgs}.
First, when applying a 3DGS model trained with one camera data to the other camera, the rendered dual camera images keep the imaging characteristics of the trained camera, as shown in \cref{fig:more_vis_comp_zoomgs}(a) and (b).
Second, when naively mixing dual-camera data to train a 3DGS model, it easily produces some visual artifacts, as shown in  \cref{fig:more_vis_comp_zoomgs}(c).
Third, by constructing 3D models for each camera, ZoomGS renders dual images that are more consistent with the GT, as shown in \cref{fig:more_vis_comp_zoomgs}(d).
Besides, we provide some examples of zoom sequences rendered from ZoomGS at \href{https://dualcamerasmoothzoom.github.io}{https://dualcamerasmoothzoom.github.io}.

\section{More Visual Results of FI models}
\label{sec:supp_3}
We provide more visual comparisons of FI models on the synthetic and real-world datasets, as shown in \cref{fig:more_syn_vis_comps} and \cref{fig:more_real_vis_comps} respectively.
The fine-tuned FI models produce more photo-realistic details and fewer visual artifacts on both datasets.
It indicates the effectiveness of the proposed data factory.

\section{Limitation}
\label{sec:supp_4}
This work is still limited in the FI model generalization between two mobile devices (\eg, apply a model trained with images from an Xiaomi mobile phone to an OPPO mobile phone). 
When the relative positions of the dual cameras on two mobile phones are greatly different, the model trained with one mobile phone data may not generalize well to the other one.
It may need to fine-tune the FI model with the data from the other mobile phone. 

\begin{figure}[t!]
    \centering
    \begin{subfigure}{0.98\textwidth}
        \ContinuedFloat
        \begin{overpic}[width=0.98\textwidth,grid=False]
        {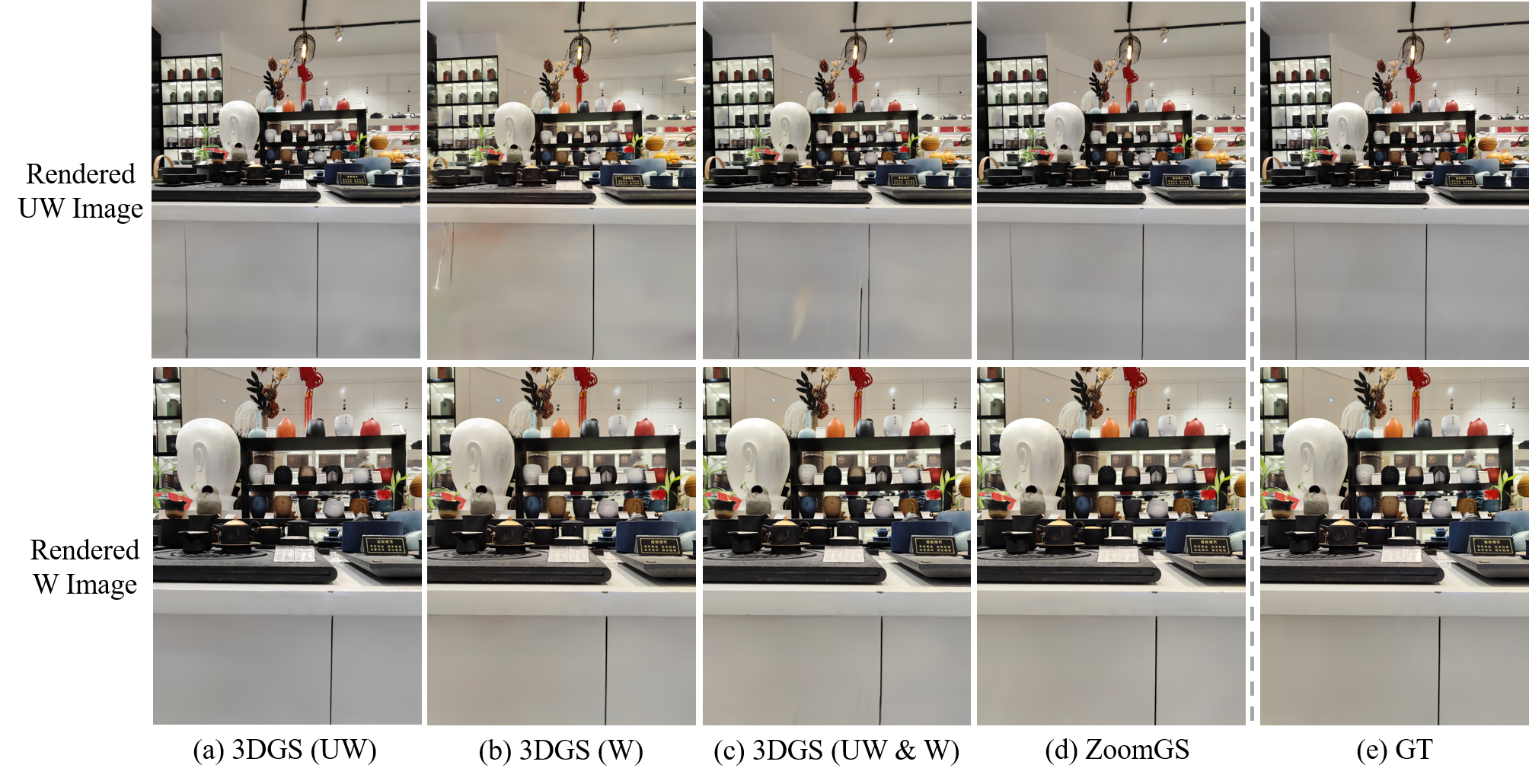}
        \end{overpic}
    \end{subfigure}
    \caption{Visual comparisons between 3DGS~\cite{3DGS} and the proposed ZommGS.
    (a) Images rendered from a 3DGS model trained with ultra-wide-angle (UW) images.
    (b) Images rendered from a 3DGS model trained with wide-angle (W) images.
    (c) Images rendered from a 3DGS model trained with UW and W images.
    (d) Images rendered from a ZoomGS model.
    (e) GT images.
    } 
    \label{fig:more_vis_comp_zoomgs}
\end{figure}

\begin{figure}[t!]
    \centering
    \begin{subfigure}{0.98\textwidth}
        \ContinuedFloat
        \begin{overpic}[width=0.98\textwidth,grid=False]
        {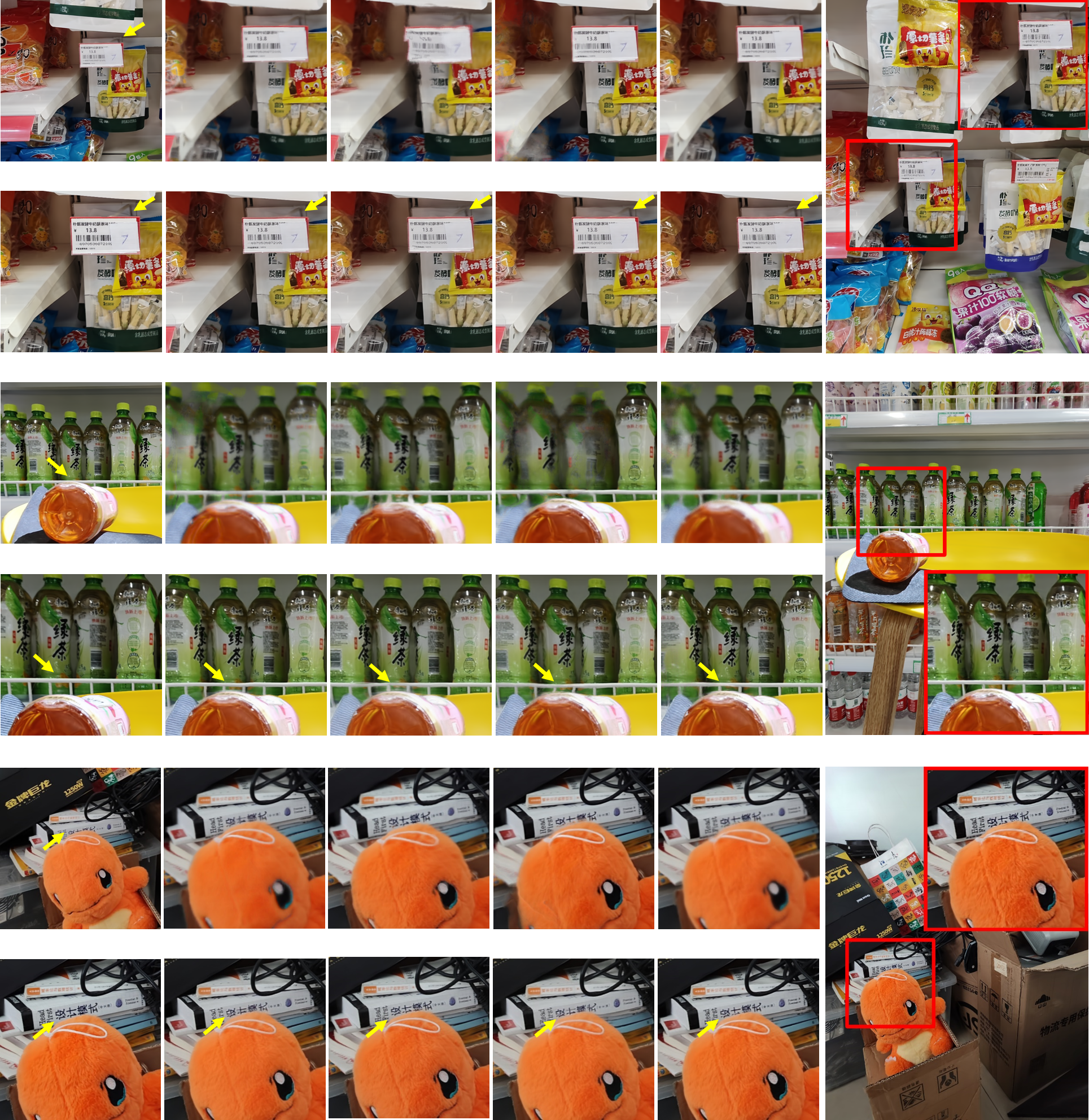}
        \put(8,52){{\fontsize{6.5}{7}\selectfont UW Image}}
         \put(59,52){{\fontsize{6.5}{7}\selectfont RIFE~\cite{RIFE}}}
         \put(110,52){{\fontsize{6.5}{7}\selectfont AMT~\cite{AMT}}}
         \put(156,52){{\fontsize{6.5}{7}\selectfont UPRNet~\cite{UPRNet}}}
         \put(205,52){{\fontsize{6.5}{7}\selectfont EMAVFI~\cite{EMAVFI}}}

         \put(8,-7){{\fontsize{6.5}{7}\selectfont W Image}}
         \put(55,-7){{\fontsize{6.5}{7}\selectfont Ours (RIFE)}}
         \put(105,-7){{\fontsize{6.5}{7}\selectfont Ours (AMT)}}
         \put(149,-7){{\fontsize{6.5}{7}\selectfont Ours (UPRNet)}}
         \put(201,-7){{\fontsize{6.5}{7}\selectfont Ours (EMAVFI)}}
          \put(283,-7){{\fontsize{6.5}{7}\selectfont GT}}

        \put(8,170){{\fontsize{6.5}{7}\selectfont UW Image}}
         \put(59,170){{\fontsize{6.5}{7}\selectfont RIFE~\cite{RIFE}}}
         \put(110,170){{\fontsize{6.5}{7}\selectfont AMT~\cite{AMT}}}
         \put(156,170){{\fontsize{6.5}{7}\selectfont UPRNet~\cite{AMT}}}
         \put(205,170){{\fontsize{6.5}{7}\selectfont EMAVFI~\cite{AMT}}}

         \put(8,111){{\fontsize{6.5}{7}\selectfont W Image}}
         \put(55,111){{\fontsize{6.5}{7}\selectfont Ours (RIFE)}}
         \put(105,111){{\fontsize{6.5}{7}\selectfont Ours (AMT)}}
         \put(149,111){{\fontsize{6.5}{7}\selectfont Ours (UPRNet)}}
         \put(201,111){{\fontsize{6.5}{7}\selectfont Ours (EMAVFI)}}
          \put(283,111){{\fontsize{6.5}{7}\selectfont GT}}

        \put(8,287){{\fontsize{6.5}{7}\selectfont UW Image}}
         \put(59,287){{\fontsize{6.5}{7}\selectfont RIFE~\cite{RIFE}}}
         \put(110,287){{\fontsize{6.5}{7}\selectfont AMT~\cite{AMT}}}
         \put(156,287){{\fontsize{6.5}{7}\selectfont UPRNet~\cite{UPRNet}}}
         \put(205,287){{\fontsize{6.5}{7}\selectfont EMAVFI~\cite{EMAVFI}}}

         \put(8,228){{\fontsize{6.5}{7}\selectfont W Image}}
         \put(55,228){{\fontsize{6.5}{7}\selectfont Ours (RIFE)}}
         \put(105,228){{\fontsize{6.5}{7}\selectfont Ours (AMT)}}
         \put(149,228){{\fontsize{6.5}{7}\selectfont Ours (UPRNet)}}
         \put(201,228){{\fontsize{6.5}{7}\selectfont Ours (EMAVFI)}}
          \put(283,228){{\fontsize{6.5}{7}\selectfont GT}}   
        \end{overpic}
    \end{subfigure}
   \caption{Visual comparisons on the synthetic dataset.
    The FI models synthesize the intermediate geometry content between dual cameras, as indicated with yellow arrows. 
    } 
    \label{fig:more_syn_vis_comps}
\end{figure}
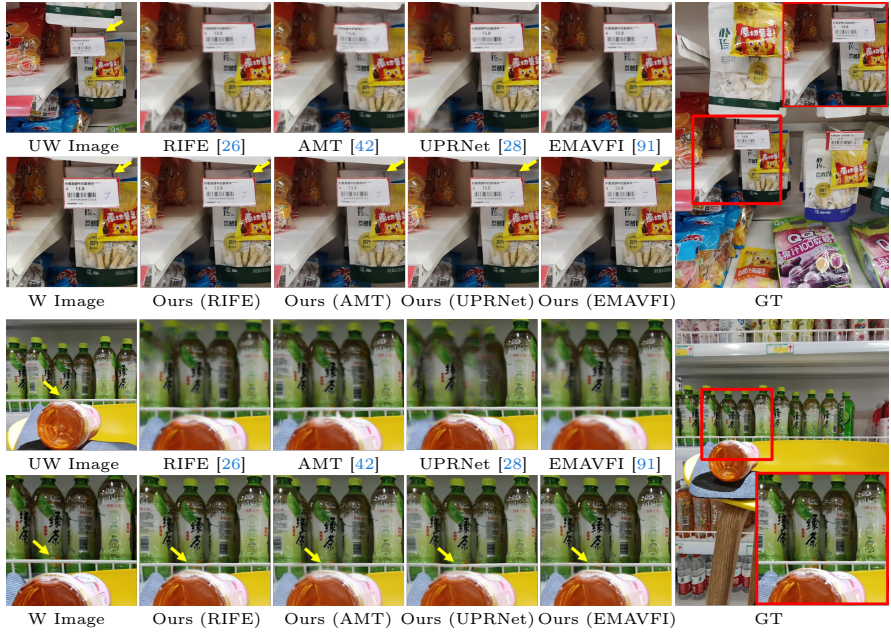

\begin{figure}[t!]
    \centering
    \begin{subfigure}{0.98\textwidth}
        \ContinuedFloat
        \begin{overpic}[width=0.98\textwidth,grid=False]
        {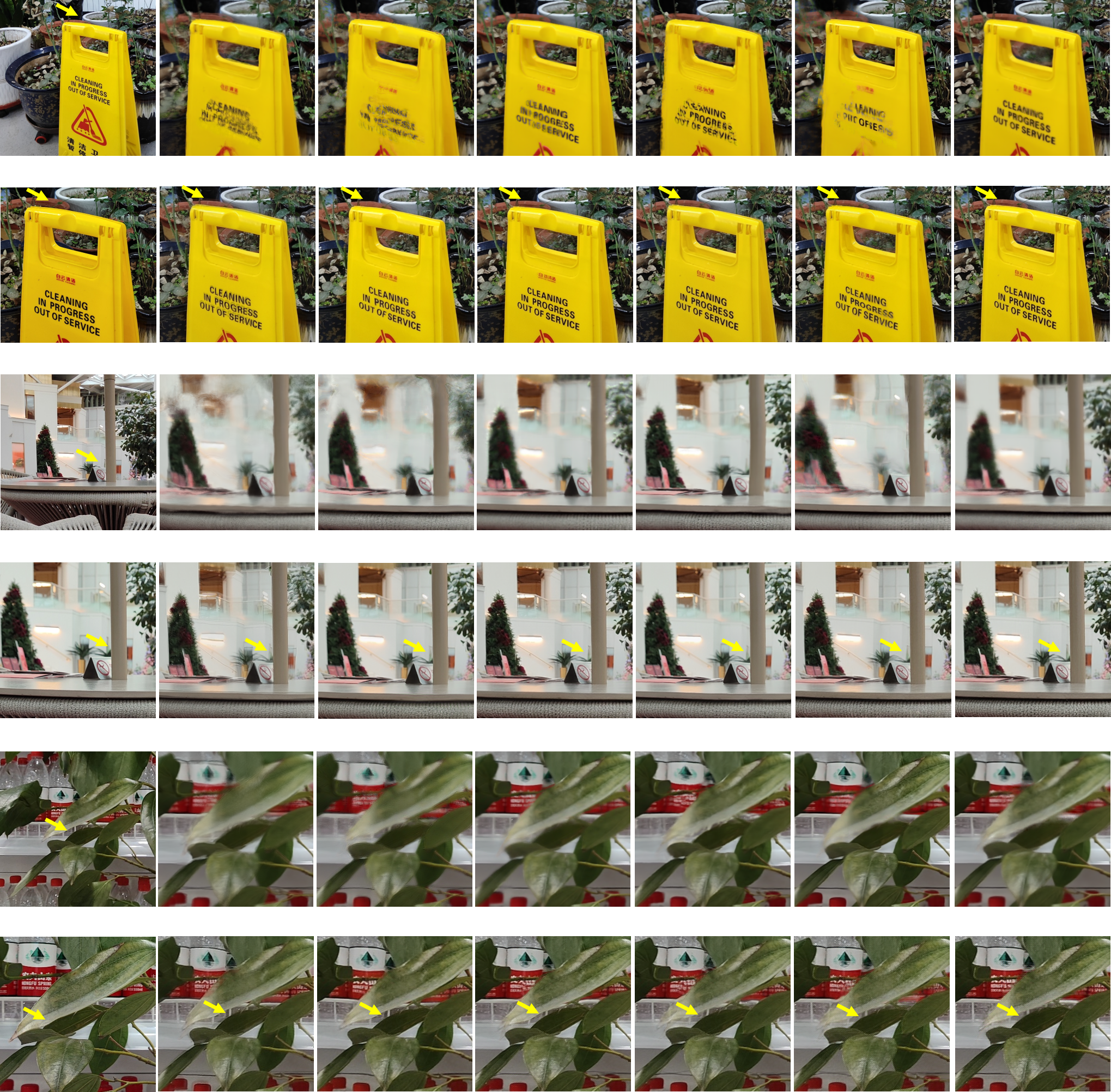}
        \put(7,50){{\fontsize{6.5}{7}\selectfont UW Image}}
        \put(55,50){{\fontsize{6.5}{7}\selectfont EDSC~\cite{EDSC}}}
        \put(100,50){{\fontsize{6.5}{7}\selectfont IFRNet~\cite{IFRNet}}}
        \put(155,50){{\fontsize{6.5}{7}\selectfont RIFE~\cite{RIFE}}}
        \put(204,50){{\fontsize{6.5}{7}\selectfont AMT~\cite{AMT}}}
        \put(247,50){{\fontsize{6.5}{7}\selectfont UPRNet~\cite{UPRNet}}}
        \put(295,50){{\fontsize{6.5}{7}\selectfont EMAVFI~\cite{EMAVFI}}}
        \put(7,-7){{\fontsize{6.5}{7}\selectfont W Image}}
        \put(43,-7){{\fontsize{6.5}{7}\selectfont Ours (EDSC)}}
        \put(91,-7){{\fontsize{6.5}{7}\selectfont Ours (IFRNet)}}
        \put(142,-7){{\fontsize{6.5}{7}\selectfont Ours (RIFE)}}
        \put(189,-7){{\fontsize{6.5}{7}\selectfont Ours (AMT)}}
        \put(231,-7){{\fontsize{6.5}{7}\selectfont Ours (UPRNet)}}
        \put(282,-7){{\fontsize{6.5}{7}\selectfont Ours (EMAVFI)}}

        \put(7,162){{\fontsize{6.5}{7}\selectfont UW Image}}
        \put(55,162){{\fontsize{6.5}{7}\selectfont EDSC~\cite{EDSC}}}
        \put(100,162){{\fontsize{6.5}{7}\selectfont IFRNet~\cite{IFRNet}}}
        \put(155,162){{\fontsize{6.5}{7}\selectfont RIFE~\cite{RIFE}}}
        \put(204,162){{\fontsize{6.5}{7}\selectfont AMT~\cite{AMT}}}
        \put(247,162){{\fontsize{6.5}{7}\selectfont UPRNet~\cite{UPRNet}}}
        \put(295,162){{\fontsize{6.5}{7}\selectfont EMAVFI~\cite{EMAVFI}}}
        \put(7,105){{\fontsize{6.5}{7}\selectfont W Image}}
        \put(43,105){{\fontsize{6.5}{7}\selectfont Ours (EDSC)}}
        \put(91,105){{\fontsize{6.5}{7}\selectfont Ours (IFRNet)}}
        \put(142,105){{\fontsize{6.5}{7}\selectfont Ours (RIFE)}}
        \put(189,105){{\fontsize{6.5}{7}\selectfont Ours (AMT)}}
        \put(231,105){{\fontsize{6.5}{7}\selectfont Ours (UPRNet)}}
        \put(282,105){{\fontsize{6.5}{7}\selectfont Ours (EMAVFI)}}

        \put(7,274){{\fontsize{6.5}{7}\selectfont UW Image}}
        \put(55,274){{\fontsize{6.5}{7}\selectfont EDSC~\cite{EDSC}}}
        \put(100,274){{\fontsize{6.5}{7}\selectfont IFRNet~\cite{IFRNet}}}
        \put(155,274){{\fontsize{6.5}{7}\selectfont RIFE~\cite{RIFE}}}
        \put(204,274){{\fontsize{6.5}{7}\selectfont AMT~\cite{AMT}}}
        \put(247,274){{\fontsize{6.5}{7}\selectfont UPRNet~\cite{UPRNet}}}
        \put(295,274){{\fontsize{6.5}{7}\selectfont EMAVFI~\cite{EMAVFI}}}
        \put(7,218){{\fontsize{6.5}{7}\selectfont W Image}}
        \put(43,218){{\fontsize{6.5}{7}\selectfont Ours (EDSC)}}
        \put(91,218){{\fontsize{6.5}{7}\selectfont Ours (IFRNet)}}
        \put(142,218){{\fontsize{6.5}{7}\selectfont Ours (RIFE)}}
        \put(189,218){{\fontsize{6.5}{7}\selectfont Ours (AMT)}}
        \put(231,218){{\fontsize{6.5}{7}\selectfont Ours (UPRNet)}}
        \put(282,218){{\fontsize{6.5}{7}\selectfont Ours (EMAVFI)}}
        \end{overpic}
    \end{subfigure}
   \caption{Visual comparisons on the real-world dataset.
    The FI models still synthesize the intermediate geometry content in the real world, as indicated with yellow arrows.  
    } 
    \label{fig:more_real_vis_comps}
\end{figure}
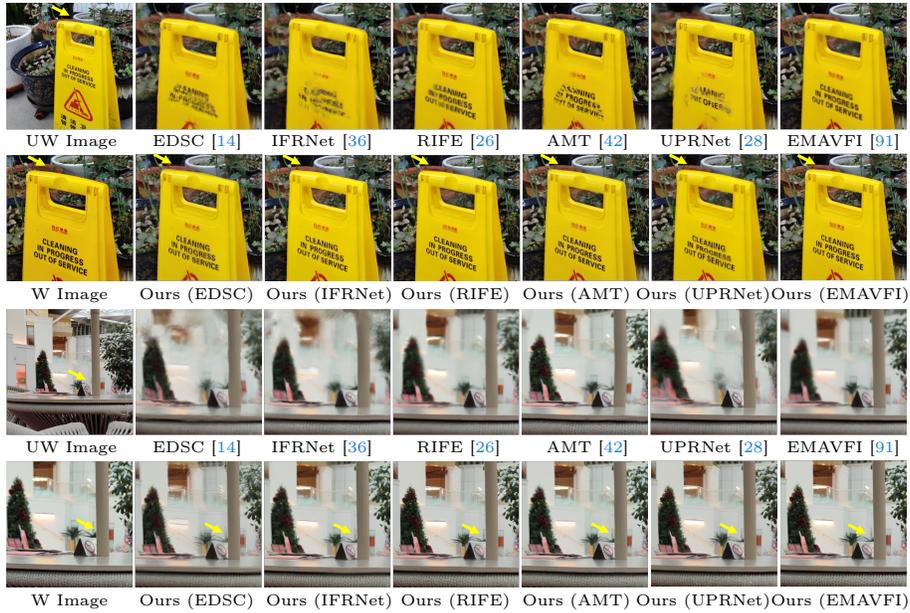

\clearpage  



%
%
\bibliographystyle{splncs04}
\bibliography{egbib}
\end{document}